%% file: iclr2027_conference.tex
\documentclass{article} 
\usepackage{iclr2027_conference,times}
\AtBeginDocument{\setlength{\headwidth}{\textwidth}}
\input{math_commands.tex}

\usepackage{hyperref}
\usepackage{url}
\usepackage{graphicx}
\usepackage{booktabs}


\title{Linear Probes to Bayesian Manifolds: Geometry of Arithmetic in Large Language Models}

\author{Anshul Kumar$^{1}$\thanks{For all correspondence, please contact Anshul Kumar at \texttt{anshul95@cmu.edu}.}, Neha Prasad$^{2}$ \\
\textbf{Manoj Kumar$^{3}$, Barnab\'as P\'oczos$^{1}$, Deeksha Varshney$^{2}$} \\
\mdseries $^{1}$Carnegie Mellon University \quad $^{2}$Indian Institute of Technology Jodhpur \quad $^{3}$Indian Institute of Technology Roorkee
}

\iclrfinalcopy
\begin{document}

\maketitle

\begin{abstract}
Linear probes are a standard method to study geometric structures, such as helices and circles, from the activations of large language models (LLMs), and such structure is often taken as evidence that the model computes with it. Probe accuracy alone cannot establish this, because it conflates four separate claims. These are (1) that the concept is decodable from the activations, (2) that they form a specific shape, (3) that the shape belongs to the probed concept rather than to a related concept the model also represents, and (4) that the output of the model depends on the
shape. We call these claims decodability, shape, ownership, and causal use, and we test them separately with a four-stage Bayesian pipeline. Stage 1 verifies that a concept is decodable and that the decoding is not an artefact. Stage 2 compares candidate shapes and selects the best one only when its evidence passes a predefined threshold. Stage 3 removes a set of related concepts, specified in advance, and tests whether the shape is owned or inherited. Stage 4 intervenes on the shape, at the final token of the prompt, to test whether the output of the model depends on it. Because every stage issues its verdict on the same data, the four answers can be compared directly. We apply the pipeline to two-digit addition and multiplication
in GPT-J 6B, Llama 3.1 8B, and Pythia 6.9B. Nearly every concept is decodable, but the decoded directions overlap substantially across concepts. Every strongly supported comparison selects a ribbon, which is a helix whose radius changes with the value, or a torus, which refines rather than contradicts the helix account of prior work. Of 1,330 cells with an ownership verdict, 19\% of shapes are owned and 81\% are inherited. Owned and inherited shapes move the output about equally often. Intervening on the shape retains approximately 61\% of the effect of intervening on the full probe while using at most five dimensions, against a median of nine for the probe. Our results indicate that a probe verdict alone overstates what a model computes with, and we propose that probe results be reported as verdicts on all four claims rather than as a single accuracy.
\end{abstract}

\section{Introduction}
\label{sec:intro}

A central goal of mechanistic interpretability is to explain the computation of large language models (LLMs) in terms of concepts that humans understand \citep{modell2025origins}. Arithmetic is a useful setting for this goal because for addition and multiplication as we can name every intermediate concept, including operand digits, carries, partial products, and column sums. Recent work finds that such concepts and others are represented geometrically, on helices, circles, and curved manifolds \citep{kantamneni2025language, liu2022towards, engels2025not, gurnee2025curve}, as Section~\ref{sec:related} surveys.

A standard method for locating such structure is the linear probe \citep{alain2016understanding}, and a high probe accuracy is often taken as evidence that the model computes with the structure. Accuracy alone cannot establish this \citep{hewitt2019designing, belinkov2022probing, elazar2021amnesic}, because it answers only the first of four separate questions. These are whether the concept can be decoded from the activations, what shape they form, whether that shape belongs to the probed concept or to a related concept the model also represents, and whether the output of the model depends on it. We call the corresponding claims \emph{decodability}, \emph{shape}, \emph{ownership}, and \emph{causal use}.

The four claims can receive different answers. Under superposition one direction can carry several features \citep{elhage2022superposition}, so a probe identifies a direction or a subspace that the model may use, ignore, or share across concepts. Every intermediate concept is also a function of the operand digits, so a probe trained on one concept can succeed by decoding another. In $23 \times 17 = 391$, our running example, the units digit of the answer, 1, is fixed by the units partial product, $3 \times 7 = 21$, which also fixes the units carry, 2. A claim that a shape belongs to one concept therefore requires that its \emph{related concepts}, those that could produce the same geometry, be removed first.

\begin{figure}[t]
\centering
\includegraphics[width=\linewidth]{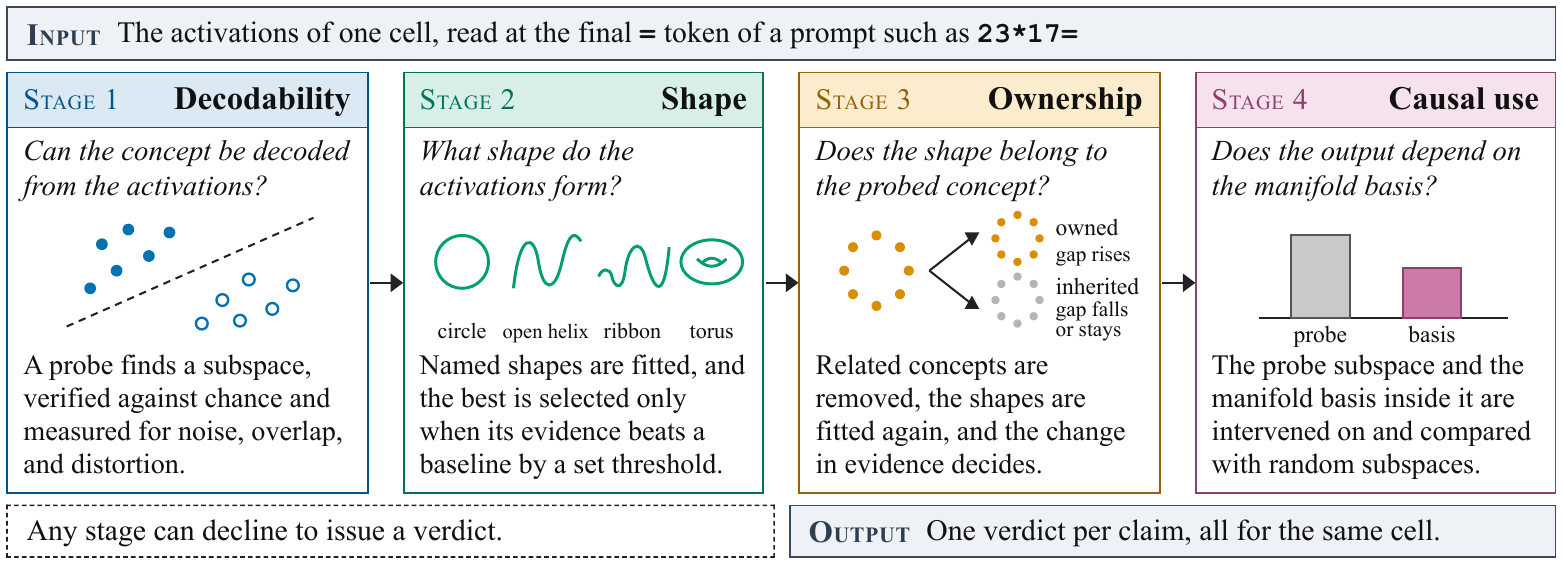}
\caption{The four-stage pipeline, one stage per claim. All four criteria are passed in only 46 of 1,330 cells.}
\label{fig:pipeline}
\vspace{-6pt}
\end{figure}

We test the four claims separately with a four-stage Bayesian pipeline (Figure~\ref{fig:pipeline}, Section~\ref{sec:method}, Appendix~\ref{app:extintro}). Stage 1 tests decodability with a probe, verifies the result against chance, and measures noise, overlap with other concepts, and distortion of distances. Stage 2 tests shape by fitting named low-dimensional manifolds, such as a circle, a torus, and a ribbon, which is a helix of varying radius. It selects the best one only when its evidence exceeds that of a generic baseline by a predefined threshold. Stage 3 tests ownership by removing the related concepts, listed before any experiment, from the activations and fitting the shapes again. A shape whose evidence over the baseline rises after the removal is \emph{owned} by the concept, and otherwise it is \emph{inherited} from a related concept. Stage 4 tests causal use by intervening on the probe subspace and on its \emph{manifold basis}, the directions inside it that the shape occupies, and compares both with random subspaces of equal rank. Bayesian model comparison scores shapes of different complexity on one scale, the marginal likelihood, and shows when no candidate is supported \citep{kass1995bayes}. Every verdict is issued for one \emph{cell}, a fixed combination of model, task, layer, concept, and preprocessing setting (Section~\ref{sec:setup}). The four verdicts therefore refer to the same data.

We limit our scope to two-digit addition and multiplication in GPT-J 6B, Llama 3.1 8B, and Pythia 6.9B \citep{wang2021gptj, grattafiori2024llama, biderman2023pythia}, and to a single token position. All activations are read, and all interventions applied, at the final \texttt{=} token of the prompt, which immediately precedes the answer and at which the answer is assembled from information moved there \citep{mamidanna2025all}. Circuit-level analyses \citep{wang2023interpretability, conmy2023towards, kramar2024atp} study the computation across operand positions and attention routes \citep{stolfo2023mechanistic}. Our results describe the geometry at the position where the answer is read out.

\begin{itemize}
\item \textbf{Decodability is nearly universal but heavily shared}. Without preprocessing, 89\% of the 1,620 cells are decodable, yet 78\% to 91\% of concept pairs align more closely than chance.
\item \textbf{Every strongly supported comparison selects a ribbon or a torus} (338 cells).
\item \textbf{Most shapes are inherited, not owned} (81\% against 19\% of the 1,330 Stage 3 verdicts).
\item \textbf{The manifold basis retains 61\% of the causal effect of the probe} (from 0.32 to 0.70).
\end{itemize}

The four claims therefore separate in practice, and only 46 of the 1,330 cells pass all four criteria. We read this count as conservative, because a cell must meet the threshold of every stage, and as conditional on the listed related concepts. The closest work fits a generalised helix and confirms it by activation patching \citep{kantamneni2025language}. We compare candidate shapes by evidence, measure the change after removing related concepts listed before any experiment, and test interventions on the directions of the shape, all on the same cells. We propose that a probe result be reported as four verdicts and not as one accuracy.

\section{Related Work}
\label{sec:related}

\paragraph{Probes and superposition.}
Linear probes \citep{alain2016understanding} read concepts from directions of the residual stream \citep{park2024linear, park2025geometry, gurnee2024language}. Control tasks \citep{hewitt2019designing} separate what the activations encode from what a probe learns. Amnesic probing \citep{elazar2021amnesic}, the survey of \citet{belinkov2022probing}, and causal readings of probe results \citep{joshi2026causality, sharma2026dissociating} separate decoding from use. Superposition explains why one direction serves several concepts \citep{elhage2022superposition}. The four claims turn these concerns into four verdicts on one cell.

\paragraph{Arithmetic in language models.}
Small transformers solve modular addition on circles, by rotation \citep{liu2022towards, nanda2023progress} or by other algorithms on the same circles \citep{zhong2023clock}. In pretrained models, addition is localised to a few attention heads and feed-forward layers \citep{stolfo2023mechanistic, zhang2024interpreting}, assembled at the last token from information moved there \citep{mamidanna2025all}, and carried in part by Fourier features \citep{zhou2024pretrained} or, in a competing account, by neuron-level heuristics \citep{nikankin2025arithmetic}. Numbers are reported on a helix \citep{kantamneni2025language}, on base 10 digit circles \citep{levy2025language}, on a line \citep{zhu2025language}, and as periodic features shared across models \citep{fu2026convergent}. The partial products of multiplication are reported as Minkowski sums of digit embeddings in a model trained with implicit chain of thought \citep{bai2025multiplication}. \citet{wen2026shape} describe the shape at the digit positions of a multi-digit answer, \citet{feucht2026arithmetic} study the same position as we do and test whether the model computes along the circle, and neither tests whether the shape is inherited.

\paragraph{Manifolds and steering.}
Concepts represented on curves include the days of the week \citep{engels2025not} and character counts \citep{gurnee2025curve}. Such curves are predicted by a continuous correspondence between a concept and its representation \citep{modell2025origins} and produced by symmetry in the statistics of language \citep{karkada2026symmetry}. \citet{gurnee2025curve} validate one manifold with interventions, \citet{li2025rema} cover many tasks but establish decodability alone, and steering along a fitted manifold \citep{wurgaft2026manifold, sarfati2026shape, modell2026probing} establishes causal use without comparing named shapes. We compare named shapes by evidence, test whether the selected shape is owned, and intervene on its manifold basis, on the same cells.

\paragraph{Concept removal and interventions.}
Projecting a concept out of the activations \citep{ravfogel2020null, belrose2023leace} and measuring what remains usable \citep{elazar2021amnesic} or how completely and selectively the removal acts \citep{canby2025reliable} are the ancestors of Stage 3, which refits the shape rather than a classifier, and of the ablation of Stage 4. Causal abstraction gives interventions their meaning \citep{geiger2025causal}, activation patching gives them their form \citep{heimersheim2024how}, and the subspace illusion \citep{makelov2024subspace} sets the limit of what a control by random subspaces of matching rank can show.

\paragraph{Probabilistic geometry.}
The shape model is the linear-Gaussian limit of the Gaussian process latent variable model \citep{lawrence2005probabilistic}. Isomap \citep{tenenbaum2000global} and uniform manifold approximation and projection \citep[UMAP,][]{mcinnes2018umap} do not supply the marginal likelihood on which the comparison rests, so they cannot say whether a shape is supported \citep[cf.][]{hauberg2018only}, which a verdict on the shape claim requires.

\section{Problem Setup and Notation}
\label{sec:setup}

\paragraph{Models and activations.}
The three models are base models without instruction tuning, share hidden size $d = 4{,}096$, and are those in which the number helix was reported \citep{kantamneni2025language}. For each model we read the residual stream at five layers of matched relative depth, which are layers 4, 8, 14, 20, and 24 of 28 in GPT-J and 4, 8, 16, 24, and 28 of 32 in Llama and Pythia, always at the final \texttt{=} token (Appendix~\ref{app:models}).

\paragraph{Tasks and data.}
Both tasks use operands from 0 to 99 and one fixed prompt that ends in the problem, such as \texttt{23*17=} (Appendix~\ref{app:data}), so every verdict is conditional on that prompt. We keep only problems whose correct answer is a single token in all three models, which holds for all 10,000 addition problems and for 3,023 multiplication problems. We analyse only the problems that a model answers correctly (Table~\ref{tab:correct} gives the shares), because our question concerns the geometry that accompanies a successful computation, and we make no claim about problems answered incorrectly.

\paragraph{Concepts.}
The single concepts are the quantities that the standard written algorithms name, together with simple properties of the operands and of the answer such as parity, magnitude, and comparisons, which gives 41 for addition and 47 for multiplication. Ten joint concepts per task, listed before any experiment and each formed from two or three single concepts, bring the totals to 51 and 57. All 108 concepts enter the decodability and shape counts that we report. Six \emph{core concepts} per task, all quantities of the standard written algorithms, are also reported separately, and we call the others \emph{extended concepts}. For addition the core concepts are the operands $a$ and $b$, the sum $a+b$, the units and tens digits of the answer, and the units carry. For multiplication they are the units and tens digits of $a$, the units digit of $b$, the units partial product, the units carry, and the units digit of the answer, which take the values 3, 2, 7, 21, 2, and 1 in the running example (Appendix~\ref{app:concepts} gives the definitions and lists all concepts). Where the arithmetic ties a concept to others, we also list these related concepts (Section~\ref{sec:intro}). The related concepts of the units digit of the answer in multiplication, for example, are the operands, their units digits, the units partial product and column sum, which are equal, and the units carry. Such a list exists for 46 of the 108 concepts, and only the cells of these concepts receive an ownership verdict.

\paragraph{Cells and notation.}
Each analysis runs under one of three preprocessing settings $u$, named off, answer, and norm. The first leaves the activations unchanged. The second removes the unembedding direction of the answer token, because towards the output the residual stream aligns with that direction, and this readout effect could be mistaken for the geometry of a concept. The third rescales every activation to unit length, because the norm of the residual stream varies widely (Appendix~\ref{app:models}) and a shape should not depend on it. The two altered settings therefore repeat every verdict without readout alignment and without activation scale, and Section~\ref{sec:results} pools the three settings from Stage 2 on. A cell is $\kappa = (M, t, \ell, c, u)$, with model $M$, task $t$, layer $\ell$, concept $c$, and setting $u$. Three models, five layers, 108 concepts, and three settings give 4,860 cells, 1,620 \emph{candidate cells} per setting. A value enters the analysis only if at least 30 correctly answered problems take it, a \emph{floor} set because smaller groups give unstable covariance estimates, and a cell only if at least two of its values remain (Appendix~\ref{app:cells}). For one cell, $\mX \in \R^{N \times d}$ stacks the activations of the $N$ remaining problems and $\vy$ stacks their concept values. We write $N_v$ for the number of problems with value $v$ and $\vmu_v$ for their mean activation, the centroid of $v$ (Appendix~\ref{app:notation} lists all symbols, terms, and abbreviations).

\section{Method: A Four-Stage Pipeline}
\label{sec:method}

The pipeline takes one cell, with its activations $\mX$ and concept values $\vy$, and issues one verdict per claim. Every later stage fits or intervenes on directions of the subspace that Stage 1 returns, so the four verdicts describe the same directions. The same thresholds apply to every cell (Appendix~\ref{app:method}, Table~\ref{tab:thresholds}), and a stage that cannot meet its threshold declines.

\paragraph{Stage 1, decodability.}
A concept is linearly decodable when the mean activation changes with its value, so we build the probe from the centroids. Let $\bar{\vmu}$ be the mean activation of the cell. We stack the centred centroids $\vmu_v - \bar{\vmu}$ as rows, each weighted by $\sqrt{N_v/N}$ to limit the influence of values with few problems, and take the singular value decomposition (SVD). We keep the leading right singular vectors whose singular values exceed the 99th percentile of their counterparts under 1,000 random permutations of $\vy$. The $r$ kept vectors are orthonormal and form $\mB \in \R^{d \times r}$, a basis of the \emph{probe subspace}. A cell is \emph{decodable} when $r \geq 1$, a verdict on whether the centroids vary more than chance and not on held-out accuracy, and its probe subspace is then \emph{verified}. A linear discriminant analysis \citep[LDA,][]{fisher1936use} inside $\mB$ then checks each direction with a second permutation test and with held-out nearest-neighbour accuracy, as diagnostics that leave $\mB$ and the verdict unchanged. In the running example the values are the digits 0 to 9, the activation for a correctly answered \texttt{23*17=} belongs to the group of value 1, and $\mB$ spans the directions that separate the ten centroids.

Three further checks remove no cell but qualify the verdict. The first compares the largest eigenvalue of the covariance of the activations outside $\mB$ with the noise level under the Mar\v{c}enko--Pastur law \citep{marchenko1967distribution}. The second marks two concepts as \emph{superposed} when the smallest principal angle between their subspaces lies more than $10^{\circ}$ below the fifth percentile of 1,000 pairs of random subspaces of the same ranks, with the false discovery rate (FDR) held at 0.05 \citep{benjamini1995controlling}, and the share of marked pairs among the concepts of the same model, layer, and setting is the superposition rate. The third is the Spearman correlation between the pairwise distances of the problems before and after projection onto $\mB$, which shows whether $\mB$ keeps the order of those distances.

\paragraph{Stage 2, shape.}
Stage 2 asks which named shape, if any, the activations form inside the probe subspace, using the coordinates $\vz_n = \mB^\top(\vx_n - \bar{\vmu}) \in \R^r$, where $\vx_n = \mX_{n,:}^\top$ is row $n$ of $\mX$ and $y_n$ its concept value. A candidate shape $K_k$ is a feature map $\bm{\phi}_k(v)$ that places the value $v$ on a curve, and the torus and the ribbon name the surfaces on which their curves lie. The circle has the features $\cos\omega v$ and $\sin\omega v$ with $\omega = 2\pi/T$, so for the units digit of the answer a period of $T = 10$ places the digits evenly around one turn, with 9 next to 0. The open helix adds a linear feature to the circle, the torus is the product of two circles with two periods, the ribbon is a helix whose radius changes linearly with the value, and a line and the concentric shape complete the six named shapes (Table~\ref{tab:shapes}). We model the coordinates as a linear image of the shape plus noise,
\begin{equation}
\vz_n = \mW^\top \bm{\phi}_k(y_n) + \bm{\epsilon}_n, \qquad \bm{\epsilon}_n \sim \gN(\vzero, \sigma_{y_n}^2 \mI), \qquad \emW_{i,j} \sim \gN(0, \alpha),
\label{eq:shape}
\end{equation}
where $\mW$ orients, scales, and can flatten the shape inside the subspace, $\alpha$ is the prior variance of its entries, and $\sigma_v$ gives each value its own spread. Step 2a proposes the period $T$ from a Fourier analysis of the centroids, and Step 2b supplies $\sigma_v$ from the covariance of the activations of each value. In Step 2c, $\mW$ is integrated out in closed form because the model is linear and Gaussian, which gives the marginal likelihood $Z_k$ of every shape, its \emph{evidence}. It rewards fit and penalises unused flexibility \citep{mackay1992bayesian}, so richer shapes need no separate penalty.

The baseline $K_0$ is a cubic polynomial in the value, with a smooth trend and no periodic feature. Let $K_{\hat{k}}$ be the named shape with the highest evidence and $g = \log Z_{\hat{k}} - \log Z_0$ its \emph{evidence gap}, in nats. The shape is \emph{selected} when $g \geq 5$, a Bayes factor of about 150 \citep{kass1995bayes}, and the bar rises to 10 when six or fewer values remain. The stage declines when no shape is selected or when the \emph{alignment} of the selected shape, its correlation with the concept values, is below 0.5. The strictest evidence grade, Tier A, adds two conditions and a permutation control (Appendix~\ref{app:method}). The held-out error of the shape may exceed that of the best alternative by at most 10\%, and, for stability, the shape must keep the highest evidence under all nine combinations of three random seeds and three values of $\alpha$. Tier A names the exact shape when the best shape leads the runner-up by the bar after a correction for the uncertainty of the period, which we call a \emph{strongly supported comparison}, and otherwise it names only the \emph{family} of the shape, one of three groups (Appendix~\ref{app:method}), when the family evidence leads the other families by the bar. A selected and aligned shape that fails either condition receives Tier B (Appendix~\ref{app:method} gives the lower grades). Stage 2 validates named shapes within one candidate set, one noise model, and one prior, and does not discover new ones.

\paragraph{Stage 3, ownership.}
Related concepts share directions under superposition, so a shape found for one concept may be produced by another. Stage 3 removes the related concepts by projection, as in linear concept removal \citep{ravfogel2020null, belrose2023leace}, and measures what remains. For a concept $c$, we orthonormalise the union of the Stage 1 bases of its related concepts, taken at the same model, layer, and setting, into $\mR$. We project this span out of every activation and recompute the coordinates, $\vz_n^{\perp} = \mB^\top(\mI - \mR\mR^\top)(\vx_n - \bar{\vmu})$. All shapes are fitted again with $\alpha$, the periods, and the noise held at their Stage 2 values and only the weights refitted. With $g^{\perp}$ the evidence gap after the removal, the verdict rests on
\begin{equation}
\Delta = g - g^{\perp}.
\label{eq:delta}
\end{equation}
The shape is \emph{owned} when $\Delta < 0$ and \emph{inherited} when $\Delta \geq 0$. The removal is linear and deletes only the span of $\mR$, so Appendix~\ref{app:method} (Equation~\ref{eq:delta_terms}) decomposes $\Delta$ into the advantage of the selected shape over the baseline on the removed component of the coordinates and the interaction between that component and the remaining one. The interaction term vanishes when $\operatorname{span}(\mR) \subseteq \operatorname{span}(\mB)$, that is, when the removal lies entirely within the probe subspace. The rule is strict at the boundary, so a cell whose coordinates the removal leaves unchanged ($\Delta = 0$) counts as inherited, as does a representation that is the sum of the representations of its related concepts. In the running example, the units digit of the answer is a function of the units partial product alone, so each of its centroids is a weighted mean of the centroids of that related concept, its probe subspace lies within their span, and an inherited verdict is the expected outcome. The verdict is operational and relative to the listed related concepts, and it does not show that the shape is unique to $c$ among all concepts (Appendix~\ref{app:method} gives its two limits).

\paragraph{Stage 4, causal use.}
The first three stages describe the activations. Stage 4 changes them and observes the output, as a causal claim requires \citep{geiger2025causal}. The interventions act on $\mB$ and on the \emph{manifold basis} $\mQ_{\mathrm{geom}}$, an orthonormal basis of the column space of $\mB\hat{\mW}^\top$, where $\hat{\mW}$ is the posterior mean of $\mW$ for the selected shape without its constant row. The manifold basis spans the directions of $\mB$ that the shape occupies, and its rank is at most five, the number of non-constant features of the ribbon. Neither intervention moves an activation along the shape, as manifold steering does \citep{wurgaft2026manifold, sarfati2026shape, modell2026probing}. Let $\vh$ be the activation at the final \texttt{=} token of layer $\ell$ and $\mP$ either basis. \emph{Ablation} replaces the part of $\vh$ inside the subspace by its mean, and \emph{patching} replaces it by that of a donor problem with a different concept value \citep{heimersheim2024how},
\begin{equation}
\vh \mapsto \vh - \mP\mP^\top(\vh - \bar{\vh}) \qquad \text{and} \qquad \vh \mapsto \vh + \mP\mP^\top(\tilde{\vh} - \vh),
\label{eq:interventions}
\end{equation}
where $\bar{\vh}$ is the mean of $\vh$ over the correctly answered problems and $\tilde{\vh}$ is the activation of the donor. In the running example, ablation removes what separates the digit 1 from the other digits, and patching inserts the digit of a donor whose answer ends differently.

The \emph{accuracy change} is the difference, in percentage points, between the shares of problems answered correctly with and without ablation. The \emph{causal excess} is the change in the logit of a target token minus the mean change under five random subspaces of the same rank. The target token is the correct answer for ablation and the donor answer for patching, so the expected excess is negative under ablation and positive under patching. The random subspaces, drawn in the full activation space, control for disturbing any subspace of that rank, and not for a pathway that is dormant on ordinary inputs \citep{makelov2024subspace}. A cell is \emph{causally active} when the causal excess of ablating $\mQ_{\mathrm{geom}}$ is at least 0.5 logits in absolute value. A change in either direction shows that the output depends on the manifold basis, so the count uses the absolute value, and Section~\ref{sec:results} reports the signs and the stricter level of 1.0. The \emph{retention} is the mean accuracy change under ablation of $\mQ_{\mathrm{geom}}$ divided by that under ablation of $\mB$, over the same cells. Stages 3 and 4 issue verdicts only for cells whose shape reaches Tier A or Tier B. A cell satisfies all four claims when it is decodable, its shape reaches Tier A and is owned, and it is causally active.

\paragraph{Fixed thresholds.}
The prior variance of the two cross features of the ribbon is scaled by 0.1, which lowers the prior weight of the cross features. Only the floor, the prior variance $\alpha$, the causal level, and the FDR of the permutation control of Tier A were also examined at other values (Section~\ref{sec:results} and Appendices~\ref{app:concepts} and~\ref{app:method}).

\section{Results}
\label{sec:results}

We ran the pipeline on 4,860 cells, 1,620 in each of three preprocessing settings. Without preprocessing 89\% of the cells are decodable, every strongly supported shape comparison ends in a ribbon or a torus, 19\% of the 1,330 ownership verdicts are owned and 81\% inherited, the manifold basis keeps 61\% of the accuracy effect of the probe subspace, and 46 cells pass all four criteria. Table~\ref{tab:results} gives every stage by model and task. The cells behind it are not independent, because one concept appears at five layers and in three settings and some concepts are functions of others, so we report counts and shares and attach no significance test to them.

\begin{table}[t]
\caption{Results by model and task. Superposition is the mean rate over the cells of the row. Ribbon, torus, and other count the cells with a shape fit, 1,457 in all, pooled over the three settings, by the shape with the highest evidence, and other covers the baseline, the circle, and the open helix. Owned and inherited count the cells with an ownership verdict, 1,330 in all. Ablation gives the mean accuracy change in percentage points under ablation of $\mB$ and of $\mQ_{\mathrm{geom}}$ over the same 1,330 cells, and retention is their ratio.}
\label{tab:results}
\centering
\small
\setlength{\tabcolsep}{3pt}
\begin{tabular}{@{}llrrrrrrrrr@{}}
\toprule
& & Stage 1 & \multicolumn{3}{c}{Stage 2, shape} & \multicolumn{2}{c}{Stage 3, ownership} & \multicolumn{3}{c@{}}{Stage 4, ablation} \\
\cmidrule(lr){3-3} \cmidrule(lr){4-6} \cmidrule(lr){7-8} \cmidrule(l){9-11}
Model & Task & Superposition & Ribbon & Torus & Other & Owned & Inherited & $\mB$ & $\mQ_{\mathrm{geom}}$ & Retention \\
\midrule
GPT-J & addition & 0.808 & 161 & 42 & 0 & 24 & 147 & $-21.5$ & $-15.0$ & 0.70 \\
GPT-J & multiplication & 0.897 & 238 & 43 & 7 & 50 & 199 & $-11.2$ & $-7.6$ & 0.68 \\
Llama & addition & 0.782 & 173 & 37 & 1 & 42 & 151 & $-6.3$ & $-2.0$ & 0.32 \\
Llama & multiplication & 0.907 & 256 & 19 & 5 & 46 & 226 & $-4.0$ & $-2.2$ & 0.55 \\
Pythia & addition & 0.815 & 141 & 61 & 3 & 42 & 149 & $-15.5$ & $-6.7$ & 0.43 \\
Pythia & multiplication & 0.909 & 239 & 28 & 3 & 49 & 205 & $-20.0$ & $-13.8$ & 0.69 \\
\midrule
All & & & 1,208 & 230 & 19 & 253 & 1,077 & $-12.6$ & $-7.7$ & 0.61 \\
\bottomrule
\end{tabular}
\end{table}

\subsection{Decodability Is Nearly Universal but Heavily Shared}

Without preprocessing, 1,445 of the 1,620 candidate cells (89\%) are decodable (Table~\ref{tab:eligibility} traces every population). The verified probe subspaces have a median rank of nine, a mean of 18.3, and a maximum of 199. The units digit of the answer is decodable in all three models.

The superposition rate averages 0.80 on addition and 0.90 on multiplication (Table~\ref{tab:results}), so the subspaces of most pairs of concepts align more closely than chance, in part because the values of related concepts are functions of the same digits, and probe accuracy alone is not evidence of structure specific to one concept. In the addition cells, where the Mar\v{c}enko--Pastur edge is a usable reference (Appendices~\ref{app:method} and~\ref{app:results}), the largest residual eigenvalue outside the probe subspace lies at 6 to 8 times the noise level on average and at up to 27 times, so the residual stream carries structure outside the probe subspace. Inside the subspace, the Spearman correlation of the pairwise distances before and after projection deviates from one by a median of $1.5 \times 10^{-4}$, so Stage 2 fits its shapes to coordinates that keep the distance ordering.

\subsection{Every Strongly Supported Comparison Selects a Ribbon or a Torus}

Among the 1,457 cells with a shape fit, pooled over the three settings, the ribbon has the highest evidence in 1,208 and the torus in 230 (Table~\ref{tab:results}). The ribbon or the torus is the selected shape in each of the 338 strongly supported comparisons, 111 of them among the core concepts. The other 19 cells fall to the baseline (13), the circle (4), and the open helix (2). For concepts that span the two-digit range, such as the operands and the sum, the ribbon mostly has the highest evidence, since its linear feature carries the trend. For residues modulo 10 the torus has it in many cells, with the exceptions listed below, and concepts of mixed range are split between the two. Among the core concepts without preprocessing, the highest evidence goes to the ribbon for some residues as well, for the units digit of $a$ in multiplication in every model, for every concept with a verdict in Llama, and for the units digits of $b$ and of the answer in GPT-J multiplication. For the operand $b$ in Pythia addition the baseline has the highest evidence, so the comparison can prefer the baseline to every named shape (Appendix~\ref{app:results}). In multiplication the units digit of the answer, our running example, gives the ribbon the highest evidence in GPT-J and Llama and the torus in Pythia.

With the weights of its two cross features, the products of its linear and circular features, at zero the ribbon is the open helix, and with the weight of its linear feature at zero as well it is the circle (Table~\ref{tab:shapes}), so the evidence selects it only when these features are supported. The torus has the highest evidence in the 230 cells where a product of two circles fits better. The open helix among the candidates is the single-period form of the helix of \citet{kantamneni2025language}, which the ribbon contains and the evidence prefers. Among the core concepts, the median evidence gap, which depends on the number of problems and of coordinates of a cell (Appendix~\ref{app:method}), is 10,866 nats for strongly supported comparisons, 224 nats for Tier A verdicts that name a family, and 509 nats for Tier B. The bar of 5 nats is therefore a necessary condition and not the deciding one.

\subsection{Most Shapes Are Inherited, Not Owned}

Of the 1,330 cells with an ownership verdict, 253 (19\%) are owned under the rule of Stage 3 and 1,077 (81\%) are inherited, and the owned share lies between 14\% and 22\% in every model and task (Table~\ref{tab:results}) and is 18\%, 20\%, and 19\% under the settings off, answer, and norm (Table~\ref{tab:bysetting}). These cells have a shape of Tier A or Tier B, a list of related concepts, and a manifold basis of positive rank, and they are also the cells eligible for intervention. Among the core concepts the inherited share is higher still, with 16 of 337 verdicts owned (5\%), and the running example gives the outcome that Section~\ref{sec:method} expects, since the units digit of the answer is inherited in all 30 of its cells with a verdict, and the tens column sum of GPT-J addition gives the other outcome at layer 20 and passes all four criteria (Appendix~\ref{app:results}, Figure~\ref{fig:geometry}).

The verdicts lie far from the boundary. Over the 1,330 cells the smallest absolute value of $\Delta$ is 1.51 nats and the median about 3,024 nats, four verdicts lie within 5 nats of zero and none at it, so a margin of 5 nats would change four verdicts of 1,330 and none among the core concepts, whose smallest is 10.58 nats. Among the core concepts the removal takes on average 0.85 of the variance of the coordinates on owned cells and 0.89 on inherited cells, and leaves the alignment on average at 0.97 and 1.03 of its value, so owned cells lose about as much variance as inherited cells and their verdict rests on the evidence gap and not on the size of the projection.

\subsection{The Manifold Basis Retains 61\% of the Causal Effect of the Probe}

Averaged over the 1,330 cells, ablating $\mB$ at the final \texttt{=} token lowers the accuracy by 12.6 points and ablating $\mQ_{\mathrm{geom}}$ by 7.7 points, a retention of 61\%, between 0.54 and 0.67 when the concepts are resampled (Appendix~\ref{app:results}). Over the cells of each model, retention is 0.69 in GPT-J, 0.60 in Pythia, and 0.42 in Llama, and by setting it is 0.64, 0.61, and 0.57 in the order of Table~\ref{tab:bysetting}, where ablating $\mB$ itself lowers the accuracy by 5.0 points against 15.4 in GPT-J and 18.1 in Pythia, over the same cells. The retention is a ratio of effects and not of ranks, and the causal excess controls for rank.

The causal excess has the expected sign. On the core concepts, the mean under ablation is $-0.85$ logits at $\mB$, with a standard deviation (SD) of 1.03, and $-0.55$ at $\mQ_{\mathrm{geom}}$ (SD 0.59), the mean under patching is $+1.70$ (SD 1.77) and $+1.02$ (SD 0.99), and every concept-level mean under ablation of $\mQ_{\mathrm{geom}}$ is negative. Of the 337 cells of the core concepts, 154 (46\%) are causally active at 0.5 logits and 66 (20\%) at 1.0, all of them with a negative excess, and under patching $\mQ_{\mathrm{geom}}$ keeps 60\% of the effect of $\mB$. For the running example, the mean causal excess under ablation of $\mQ_{\mathrm{geom}}$ is $-0.78$ logits in GPT-J, $-0.40$ in Llama, and $-0.82$ in Pythia.

\begin{figure}[t]
\centering
\includegraphics[width=0.9\linewidth]{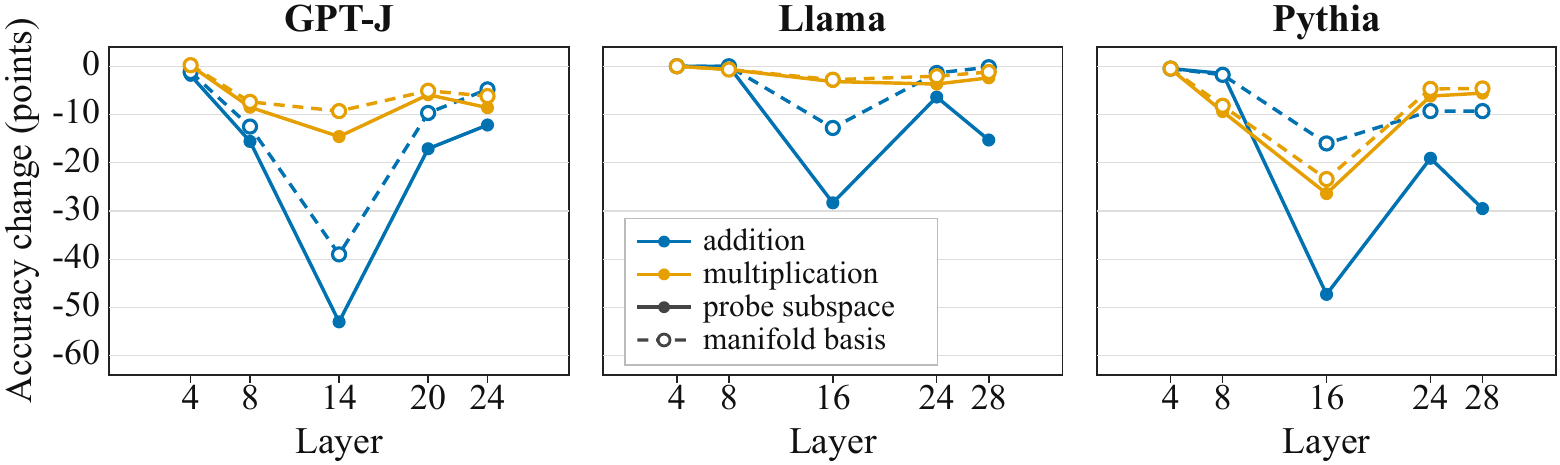}
\caption{Mean accuracy change under ablation of the probe subspace $\mB$ (solid) and of the manifold basis $\mQ_{\mathrm{geom}}$ (dashed) by layer, over the core concepts and the three settings of each model and task.}
\label{fig:depth}
\vspace{-6pt}
\end{figure}

The largest effect of ablating $\mB$ lies at the middle layer, 14 of 28 or 16 of 32, in five of the six model and task pairs and at layer 24 in Llama multiplication, and at layer 4 every change lies within two points of zero (Figure~\ref{fig:depth}). A geometry that served only the readout, where the residual stream aligns with the unembedding direction \citep{belrose2023tuned}, would matter most at the deepest layers, so the profile does not match that of a readout direction alone.

\paragraph{All four claims.}
All four criteria are passed in 46 of the 1,330 cells, 38 with a strongly supported comparison and 8 with a Tier A verdict that names only the family. They lie in every model and task, 37 with a ribbon and 9 with a torus, and 26 combinations of model, task, layer, and concept account for them, 15 of which pass in at least two of the three settings and 5 in all three. Causal activity is about as common among owned shapes as among inherited ones, 103 of the 253 owned cells (41\%) against 42\% of the inherited cells, so an ownership verdict does not stand in for a causal one. Which concepts own their shapes follows the direction of the arithmetic, since the answer digits and the carries own none of their 217 cells, while the operands, their digits and functions, the column and running sums, the partial products, and the sum own 253 of 1,113, and all 46 cells are of these kinds (Table~\ref{tab:bykind}). The units digit of the answer shows the separation. It is decodable, forms a ribbon or a torus, inherits that shape as its dependence on the units partial product predicts, and passes the causal level on average in two of the three models, so probe accuracy alone would have credited it with a shape of its own.

\section{Analysis}
\label{sec:analysis}

\paragraph{Ribbons and tori against the circles and helices of prior work.}
Two differences from the experiments of prior work \citep{liu2022towards, nanda2023progress, zhong2023clock, kantamneni2025language} bear on the shapes that Stage 2 selects. The arithmetic on the operands 0 to 99 is not modular, so a two-digit quantity advances along a line while its units digit repeats, which the linear and periodic features of the ribbon carry. A residue modulo 10 advances along no line, and for these concepts the torus, a product of two periods, has the highest evidence more often than the circle. By the final \texttt{=} token the residual stream has begun to turn towards the unembedding direction of the answer, a component that the linear and cross features of the ribbon carry and the circle cannot, which is a candidate explanation that a separate experiment would test.

\paragraph{Depth.}
The largest effects of ablating a manifold basis lie at the middle layers, and owned shapes appear at the two earliest layers in every model and task, 88 of the 253 in all, so neither the effect of a manifold basis nor the ownership of its shape is confined to the layers at which the residual stream turns towards the answer.

\paragraph{Structure outside the probe subspace.}
The structure outside the probe subspace and the overlap of the probe subspaces describe one state of the residual stream at the final token, in which many concepts share directions, so the shape found in one probe subspace can belong to another concept.

\paragraph{What the four verdicts add to a probe.}
Most shapes that a probe recovers are inherited under the rule of Stage 3, the quantities that enter a column of the written algorithm own shapes and the digits and carries that leave one never do, and owned shapes are causally active no more often than inherited ones. A probe accuracy therefore credits a concept with a shape it may inherit, an ownership verdict credits it with a shape it may not use, and an intervention on the probe subspace credits it with an effect that the manifold basis carries only in part.

\paragraph{Robustness and scope of the interventions.}
The cells of the core concepts that lose values stay verified at floors of 20 and 50, the ribbon and the torus remain the dominant shapes of Tier A at FDR levels of 0.05 and 0.20 of its permutation control, and every Tier A shape is stable over three values of the prior variance by construction (Appendices~\ref{app:concepts} and~\ref{app:method}). Each intervention on the manifold basis is compared with random subspaces of the same rank, so the causal excess reports what the directions of the shape add beyond a perturbation of that rank, while the distance of the changed activation from the native distribution \citep{heimersheim2024how, hanna2024faith} lies outside the scope of that control (Appendix~\ref{app:method}). Stages 1, 2, and 4 apply to any task with a labelled concept and a candidate set of shapes, and Stage 3 to any task that lists related concepts.

\section{Discussion and Conclusion}
\label{sec:discussion}

The cells that pass all four criteria are the yield of a standard that asks four questions of one cell. Every cell with a verdict that falls short does so for a stated reason, no candidate shape at Tier A, a shape that the listed related concepts account for, or a shape whose intervention effect stays below the causal level. The causal measures concern the logit and the accuracy of the answer token, so a shape may serve other outputs without appearing in these measures. Where all four criteria are passed, the interventions show that changing the directions of the shape moves the output, and necessity would need interventions that remove every substrate of the concept at once. The three model families cover two block layouts and three tokenisers at 6 to 9 billion parameters, with the owned share from 14\% to 22\% and the retention between 0.32 and 0.70 across the six pairs of model and task.

A geometric claim about an arithmetic representation then has four parts, a subspace of the residual stream, a curve inside it, the concept that owns the curve, and an effect on the output, and a probe accuracy settles only the first. The manifold basis may serve as a handle for model editing, and arithmetic beyond single-token answers is the next task.

\clearpage

\subsection*{AI use statement}
In this work, we used generative AI tools to aid or polish the writing. We have reviewed all AI-assisted text. We take responsibility for the final content of this work, including text, claims or artifacts produced with the aid of generative AI.

\subsection*{Ethics statement}
This work analyses three publicly released base models on synthetic arithmetic problems. It involves no human subjects, uses no data of human origin, releases no model, and changes no model weights. We identified no risk beyond those documented for the models themselves.

\subsection*{Reproducibility statement}
Appendix~\ref{app:models} describes the three models and the capture convention, Appendix~\ref{app:data} the prompt and the data, Appendix~\ref{app:concepts} every concept with its definition and the lists of related concepts, Appendix~\ref{app:cells} the cell counts and the exclusion rules, Appendix~\ref{app:method} the method with every threshold (Table~\ref{tab:thresholds}), and Appendix~\ref{app:results} the values behind Figure~\ref{fig:depth}. The problems of both tasks, the definitions of the 108 concepts, the code of every stage, and the per-cell verdicts will be released with the final version of this paper.

\bibliography{iclr2027_conference}
\bibliographystyle{iclr2027_conference}

\appendix
\section{Notation and Terms}
\label{app:notation}

We follow the notation of \citet{goodfellow2016deep}, in which scalars are italic, vectors are bold lower case, and matrices are bold upper case. Tables~\ref{tab:symbols} and~\ref{tab:symbolsmethod} give the symbols of the main text with their values in the running example $23 \times 17 = 391$ where one exists, Table~\ref{tab:symbolsapp} gives the symbols introduced in Appendix~\ref{app:method}, Table~\ref{tab:abbreviations} gives every abbreviation, and Table~\ref{tab:terms} gives every term that we introduce.

\begin{table}[htp]
\caption{Symbols of Sections~\ref{sec:intro} and~\ref{sec:setup} and Appendix~\ref{app:concepts}.}
\label{tab:symbols}
\centering
\small
\begin{tabular}{@{}p{0.20\linewidth}p{0.55\linewidth}p{0.17\linewidth}@{}}
\toprule
Symbol & Meaning & Example \\
\midrule
$a$, $b$ & The two operands, integers from 0 to 99 & 23, 17 \\
$a+b$, $ab$ & The sum and the product, which are the answers in addition and in multiplication & product 391 \\
$a_{\mathrm{units}}$, $a_{\mathrm{tens}}$ & Units and tens digits of $a$, equal to $a \bmod 10$ and $\lfloor a/10 \rfloor$. The same subscripts apply to $b$, and the tens digit of an answer $x$ is $\lfloor x/10 \rfloor \bmod 10$ & 3, 2 \\
$\1_{\mathrm{condition}}$ & Indicator, equal to 1 when the condition holds and 0 otherwise & \\
$\R$ & The real numbers & \\
$M$ & A model, one of GPT-J 6B, Llama 3.1 8B, and Pythia 6.9B & \\
$t$ & A task, addition or multiplication & multiplication \\
$\ell$ & A layer of the model & \\
$c$ & A concept & units digit of the answer \\
$u$ & A preprocessing setting, one of off, answer, and norm & \\
$\kappa = (M, t, \ell, c, u)$ & A cell, the unit of analysis & \\
$d$ & Hidden size of the model, 4,096 & \\
$N$ & Number of problems in a cell, those answered correctly whose value passes the floor & \\
$\mX \in \R^{N \times d}$ & Activations of a cell, one row per problem & \\
$\vy$ & Concept values of a cell, one entry per problem & entry 1 \\
$v$, $v'$ & One value of a concept, and a second value & 1 \\
$N_v$ & Number of problems in a cell with value $v$ & \\
$\vmu_v$ & Centroid of value $v$, the mean activation of its problems & \\
\bottomrule
\end{tabular}
\end{table}

\begin{table}[htp]
\caption{Symbols of Section~\ref{sec:method} and Table~\ref{tab:shapes}.}
\label{tab:symbolsmethod}
\centering
\small
\begin{tabular}{@{}p{0.20\linewidth}p{0.55\linewidth}p{0.17\linewidth}@{}}
\toprule
Symbol & Meaning & Example \\
\midrule
$\bar{\vmu}$ & Mean activation of the $N$ problems of a cell & \\
$\mI$, $\vzero$, $\vone$, $^\top$ & The identity matrix, the vectors of zeros and of ones, and the transpose & \\
$\gN(\vmu, \mSigma)$ & Gaussian distribution with mean $\vmu$ and covariance $\mSigma$. Its density at $\vx$ is $\gN(\vx ; \vmu, \mSigma)$ & \\
$\vx_n$, $y_n$ & Activation and concept value of problem $n$ & $y_n = 1$ \\
$r$ & Rank of the probe subspace, the number of singular vectors kept in Stage 1 & \\
$\mB \in \R^{d \times r}$ & Orthonormal basis of the probe subspace found in Stage 1 & \\
$\vz_n$, $\vz_n^{\perp}$ & Coordinates of problem $n$ inside $\mB$, before and after the related concepts are removed & \\
$k$, $K_k$ & Index of a candidate shape, from 0 to 6, and the shape itself & \\
$K_0$ & Generic baseline shape & \\
$K_1, \dots, K_6$ & Named shapes, in order line, circle, open helix, torus, concentric, and ribbon & \\
$\bm{\phi}_k(v)$, $m_k$ & Feature map of the shape $K_k$, and the number of its features with the constant & \\
$s$ & A value rescaled linearly to the unit interval & \\
$T$, $\omega$ & Period of a periodic shape and its angular frequency $2\pi/T$ & $T = 10$ \\
$\mW \in \R^{m_k \times r}$, $\emW_{i,j}$, $\hat{\mW}$ & Weights that place a shape inside $\mB$, one of their entries, and their posterior mean & \\
$\bm{\epsilon}_n$ & Noise added to the coordinates of problem $n$ & \\
$\alpha$ & Prior variance of the entries of $\mW$ & \\
$\sigma_v$ & Spread of value $v$, the standard deviation of its noise, supplied by Step 2b & \\
$Z_k$ & Marginal likelihood of the shape $K_k$, also called its evidence, a scalar & \\
nat & Unit of log evidence, from the natural logarithm & \\
$K_{\hat{k}}$ & Named shape with the highest evidence & \\
$g$, $g^{\perp}$ & Evidence gap over $K_0$, before and after the removal & \\
$\mR$ & Orthonormal basis of the span of the subspaces of the related concepts & \\
$\Delta$ & The difference $g - g^{\perp}$, negative when the shape is owned & \\
$\mQ_{\mathrm{geom}}$ & Manifold basis, an orthonormal basis of the directions of $\mB$ that the selected shape occupies, of rank at most five & \\
$\mP$ & Basis on which an intervention acts, either $\mB$ or $\mQ_{\mathrm{geom}}$ & \\
$\vh$, $\bar{\vh}$, $\tilde{\vh}$ & Activation at the final \texttt{=} token of a layer, its mean, and that of a donor & \\
\bottomrule
\end{tabular}
\end{table}

\begin{table}[htp]
\caption{Symbols introduced in Appendix~\ref{app:method}.}
\label{tab:symbolsapp}
\centering
\small
\begin{tabular}{@{}p{0.20\linewidth}p{0.55\linewidth}p{0.17\linewidth}@{}}
\toprule
Symbol & Meaning & Example \\
\midrule
$V$ & Number of values of a concept that remain in a cell & 10 \\
$\mC$ & Weighted and centred centroids, one row per value & \\
$\lambda_j$, $\tau_j$ & Squared singular value of rank $j$ of $\mC$, and the percentile of the same quantity under permutation & \\
$\gamma$, $\sigma_{\mathrm{res}}^2$ & Ratio of the residual dimension to $N$, and trace of the residual covariance over the residual dimension, in the Mar\v{c}enko--Pastur check & \\
$\mX^{\mathrm{res}}$, $\lambda_1^{\mathrm{res}}$ & Parts of the activations outside $\mB$, and the largest eigenvalue of their covariance & \\
$\lambda_{\mathrm{MP}}$ & Mar\v{c}enko--Pastur edge, the largest eigenvalue that noise alone produces outside $\mB$ & \\
$\mB_c$, $\mB_{c'}$, $\theta_1$ & Probe bases of two concepts $c$ and $c'$, and the smallest principal angle between them & \\
$|| \cdot ||$, $|| \cdot ||_2$ & Euclidean norm of a vector, and largest singular value of a matrix & \\
$f$ & Frequency of the periodogram of Step 2a, in cycles per $V$ ordered values & \\
$T^{\star}$ & Period accepted by Step 2a, which seeds the candidate periods of Step 2c & \\
$\hat{\mSigma}_v$ & Estimated covariance of the activations of value $v$ inside $\mB$ & \\
$\eta(v)$ & Radius of the concentric shape at the value $v$ & \\
$\omega_1$, $\omega_2$ & Angular frequencies of the two circles of the torus & \\
$\mZ$, $\mZ^{\perp} \in \R^{N \times r}$ & Coordinates of all problems of a cell, one row per problem, before and after the removal & \\
$\mPhi_k \in \R^{N \times m_k}$ & Features of all problems of a cell under the shape $K_k$ & \\
$\mD$ & Diagonal matrix of the noise variances $\sigma_{y_n}^2$, one per problem & \\
$\mA$ & The $m_k \times m_k$ matrix from which the evidence and $\hat{\mW}$ follow & \\
$\rho$, $\rho_{\mathrm{circ}}$, $\rho_{\max}$ & Linear and circular correlations between a fitted shape and the concept values, and the larger of their absolute values, the alignment & \\
$\hat{\vw}_s$ & Row of $\hat{\mW}$ that belongs to the linear feature $s$ & \\
$\mG_k$ & Covariance $\alpha\mPhi_k\mPhi_k^\top + \mD$ of one column of coordinates under the shape $K_k$ & \\
$\mE$, $\mA_{\Delta}$ & The part of the coordinates that the removal deletes, $\mZ - \mZ^{\perp}$, and the matrix $\mG_0^{-1} - \mG_{\hat{k}}^{-1}$ of Equation~\ref{eq:delta_terms} & \\
$\ell_m$, $\delta_m$, $h_m$ & Log evidence at the candidate period $T_m$, the half distance between its two neighbouring periods, and the second difference of $\ell_m$ there & \\
\bottomrule
\end{tabular}
\end{table}

\begin{table}[htp]
\caption{Abbreviations.}
\label{tab:abbreviations}
\centering
\small
\begin{tabular}{@{}p{0.27\linewidth}p{0.67\linewidth}@{}}
\toprule
Abbreviation & Full term \\
\midrule
BPE & Byte pair encoding, the tokenisation scheme of the three models \\
CPU & Central processing unit \\
FDR & False discovery rate \\
LDA & Linear discriminant analysis \\
LLM & Large language model \\
OAS & Oracle approximating shrinkage, a covariance estimator \\
SD & Standard deviation \\
SVD & Singular value decomposition \\
UMAP & Uniform manifold approximation and projection, an embedding method \\
\bottomrule
\end{tabular}
\end{table}

\begin{table}[htp]
\caption{Terms.}
\label{tab:terms}
\centering
\small
\begin{tabular}{@{}p{0.24\linewidth}p{0.72\linewidth}@{}}
\toprule
Term & Meaning \\
\midrule
Cell & One combination of model, task, layer, concept, and preprocessing setting \\
Candidate cell & A cell of one setting before any exclusion rule applies \\
Cell with a shape fit & A cell in which Step 2c computes the evidence of every candidate shape \\
Core concepts & The six concepts per task that are also reported separately \\
Extended concepts & All other listed concepts \\
Joint concept & A concept formed from two or three single concepts \\
Related concepts & The concepts that the arithmetic ties to a concept, listed in advance \\
Floor & The minimum of 30 correctly answered problems that a value needs \\
Flagged cell & A cell with more than three times as many problems at its most frequent value as at its least frequent, kept in the analysis \\
\raggedright Decodability, shape, ownership, causal use & The four claims, tested by Stages 1 to 4 in that order \\
Decodable & A cell in which Stage 1 keeps at least one direction, $r \geq 1$ \\
Probe subspace & The span of the singular vectors that Stage 1 keeps, with orthonormal basis $\mB$ \\
Verified probe subspace & The probe subspace of a decodable cell. The three further checks of Stage 1 qualify it and remove no cell \\
Superposed & Two concepts whose subspaces align more closely than random subspaces of the same ranks \\
Superposition rate & Share of concept pairs whose subspaces align more closely than chance \\
Candidate shape & One of the seven shapes $K_0$ to $K_6$ that Stage 2 compares \\
Named shape & One of the six candidate shapes $K_1$ to $K_6$ \\
Generic baseline & The shape $K_0$, a cubic polynomial in the value with no periodic feature \\
Evidence & The marginal likelihood $Z_k$ of a shape, with the weights integrated out \\
Evidence gap & Log evidence of the best named shape minus that of the generic baseline, in nats \\
Selected & A shape whose evidence gap reaches the predefined bar of 5 nats, or 10 nats for six or fewer values \\
Alignment & The larger of the absolute linear and circular correlations between a fitted shape and the concept values \\
Decline & A stage issues no verdict for a cell because a predefined threshold is not met \\
Tier A & The strictest evidence grade for a selected shape. It names the exact shape or only its family \\
Held-out condition & The condition of Tier A that the held-out error of the shape exceeds that of the best alternative by at most 10\% \\
Stability condition & The condition of Tier A that the shape keeps the highest evidence under all nine combinations of three seeds and three values of $\alpha$ \\
Tiers B to D & Lower evidence grades. Tier B meets the evidence bar and alignment only, Tier C supports only a dimension, and Tier D is a declined verdict \\
\raggedright Strongly supported comparison & A shape comparison that reaches Tier A and names the exact shape \\
Family & One of three groups of candidate shapes, the circle and the open helix, the torus, the concentric shape, and the ribbon, and the line with the baseline \\
Owned & A shape with ${\Delta < 0}$, whose evidence gap grows when its related concepts are removed \\
Inherited & A shape with ${\Delta \geq 0}$, whose evidence gap does not grow when its related concepts are removed \\
Manifold basis & An orthonormal basis of the directions of $\mB$ that the selected shape occupies \\
Cross features & The two features of the ribbon that multiply the linear feature by the cosine and the sine \\
Ablation & Replacing the part of an activation inside a subspace by its mean \\
Patching & Replacing the part of an activation inside a subspace by that of a donor \\
Donor & A problem with another concept value whose activation is patched in \\
Accuracy change & Change in the share of correctly answered problems under ablation, in percentage points \\
Causal excess & Logit change under an intervention minus the mean change under random subspaces of the same rank \\
Causally active & A cell whose causal excess under ablation of $\mQ_{\mathrm{geom}}$ is at least 0.5 logits in absolute value \\
Retention & Mean accuracy change under ablation of $\mQ_{\mathrm{geom}}$ divided by that under ablation of $\mB$ \\
Preprocessing setting & One of the three treatments of the activations, off, answer, and norm \\
\bottomrule
\end{tabular}
\end{table}

\clearpage

\section{Extended Introduction Material}
\label{app:extintro}

\paragraph{Geometric structure in prior work.}
In three mid-sized LLMs, integer tokens lie on a generalised helix with periods 2, 5, 10, and 100, fitted at the operand tokens and, for the sum, at the last token, and activation patching confirms that the helix is causally implicated in addition and subtraction \citep{kantamneni2025language}. In small transformers trained on modular addition, small changes to hyperparameters and initialisation lead to qualitatively different solutions \citep{zhong2023clock}. Character counts in Claude 3.5 Haiku lie on low-dimensional curved manifolds that the model twists to estimate the distance to a line boundary \citep{gurnee2025curve}. On the theoretical side, \citet{modell2025origins} model a concept as a metric space. They show that, if the correspondence between concept and representation is continuous and one-to-one, the representation forms a manifold with the same topology as the concept. If cosine similarity also decreases locally with distance in the concept, the lengths of paths along the manifold recover the distances of the concept. The linear representation hypothesis holds that concepts occupy linear directions \citep{park2024linear}. Recent work still labels distance statistics over nearest neighbours a manifold without recovering geodesics, curvature, or topology \citep{li2025rema}.

\paragraph{Dependencies among concepts.}
Dependencies like the one in the running example of Section~\ref{sec:intro} hold by construction throughout the standard written algorithms, and some are exact identities (Appendix~\ref{app:concepts}).

\paragraph{Design of the pipeline.}
A subspace shows where a concept is encoded, and the shape inside it is the candidate for the structure that the model computes with, which is why Stage 2 follows Stage 1. The checks of Stage 1 follow the control-task rationale of \citet{hewitt2019designing}.

\paragraph{Scope of the single-position design.}
Every intervention overwrites the final \texttt{=} token alone, before the residual stream is passed to the next block. The pipeline does not identify which neurons, Fourier components, attention edges, or sparse-autoencoder features realise a shape, and the Fourier-feature and neuron-heuristic accounts of Section~\ref{sec:related} are candidate substrates for the same geometry.

\section{Models and Activation Capture}
\label{app:models}

The three tokenisers are variants of byte pair encoding (BPE). GPT-J uses GPT-2 BPE, Llama uses TikToken BPE, and Pythia uses GPT-NeoX BPE. GPT-J and Pythia run attention and the feed-forward layer in parallel, and Llama runs them in sequence. We exclude instruction-tuned variants because instruction tuning changes how a model follows a prompt and would confound the analysis. We stop before the final layers, where the activation at the last token is expected to align mostly with the unembedding direction of the answer token, which reflects readout rather than computation. At each layer we read the residual stream after the transformer block, with prompts padded on the left. This vector is the full state passed to the next block, not the contribution of one block alone, and the convention is the same for both block layouts. The mean norm of the residual stream grows over the five layers by a factor of 3.1 in GPT-J, 8.8 in Llama, and 3.7 in Pythia, and at each layer it is 4.8 to 20.1 times larger in GPT-J and Pythia than in Llama.

\section{Prompts, Tokenisation, and Data}
\label{app:data}

Every problem uses the prompt \texttt{Output ONLY a number. \{a\}+\{b\}=} for addition and the same with \texttt{*} for multiplication. Ending the prompt with \texttt{=} fixes the next-token position to where the answer is expected and gives every problem the same final position. We add no trailing newline, because a newline would become the final token in place of \texttt{=}. Without the instruction, GPT-J and Pythia often continue the prompt with further text, and the most probable next token can be punctuation and not the answer. Decoding is greedy, which makes the selection of correctly answered problems deterministic. A problem is usable for a model when its correct answer is a single token under the tokeniser of that model. Every integer up to 520 is a single token for GPT-J, up to 999 for Llama, and up to 530 for Pythia, whether the number follows an empty prefix, a plus sign, a multiplication sign, or an equals sign. Every addition answer lies within all three limits. Some larger products are also single tokens, and the multiplication set consists of the 3,023 pairs whose product is a single token for all three models. A problem is answered correctly when the most probable next token equals the answer token (Table~\ref{tab:correct}). Because every answer is a single token, this equals correctness of the full answer.

\begin{table}[ht]
\caption{Problems answered correctly by each model. Percentages are of 10,000 addition problems and 3,023 multiplication problems.}
\label{tab:correct}
\centering
\begin{tabular}{lcc}
\toprule
Model & Addition & Multiplication \\
\midrule
GPT-J 6B & 8,415 (84.15\%) & 2,751 (91.00\%) \\
Llama 3.1 8B & 9,963 (99.63\%) & 2,927 (96.82\%) \\
Pythia 6.9B & 7,718 (77.18\%) & 2,757 (91.20\%) \\
\bottomrule
\end{tabular}
\end{table}

\section{Concept Definitions and Related Concepts}
\label{app:concepts}

\paragraph{Definitions.}
For an operand $a$ we write $a_{\mathrm{units}} = a \bmod 10$ and $a_{\mathrm{tens}} = \lfloor a/10 \rfloor$, and likewise for $b$. An answer $x$ can exceed 99, so its tens digit is $\lfloor x/10 \rfloor \bmod 10$. The units partial product is $a_{\mathrm{units}} b_{\mathrm{units}}$, and the units column sum is the sum of the partial products in the units column, which in two-digit multiplication is the same quantity. The units carry is $\1_{a_{\mathrm{units}} + b_{\mathrm{units}} \geq 10}$ in addition and $\lfloor a_{\mathrm{units}} b_{\mathrm{units}} / 10 \rfloor$ in multiplication, so the two tasks share the name but not the quantity. The column sum of a column is the sum of the digits of the operands in that column in addition, and the sum of the partial products that land in that column in multiplication, without the carry that arrives from the column to its right. The running sum of a column is its column sum plus that incoming carry, and the carry out of a column is the running sum divided by 10 and rounded down, which for addition is 1 when the running sum reaches 10. The number of digits of a quantity is its number of decimal digits, one for zero. The magnitude tier of a quantity is zero, one digit, two digits, or three digits. The core concepts of multiplication include the tens digit of $a$ and not that of $b$, which has the same distribution by construction and is an extended concept. The ten joint concepts per task consist of four pairs of operand digits, two that bind a carry to another quantity, and four that join two operand digits with a third quantity, one of which is determined by construction. The values of a joint concept are coded in mixed radix with the first concept as the most significant part, so that \texttt{a\_units\_\_b\_units} takes the value $10\,a_{\mathrm{units}} + b_{\mathrm{units}}$ and a concept of three parts nests in the same way. The core concepts are single concepts, so the results reported for them do not depend on this order.

\paragraph{Values removed by the floor.}
Among the core concepts, the sum $a+b$ has 199 values and loses 58 to 78 of them, depending on the model. The units partial product has 37 values and loses 3 or 4. The value 8 of the units carry in multiplication is removed for all three models, with 8, 14, and 11 problems for GPT-J, Llama, and Pythia. The other core concepts keep every value. With a floor of 20 these losses shrink to 39 to 54, 1 or 2, and 1, and with a floor of 50 they grow to 98 to 125, 14 or 15, and 2, while Pythia also loses 1 value of $a$ and 5 of $b$. The probe subspaces of the cells of the core concepts that lose values remain verified under all three floors, and we report the later stages at a floor of 30 only.

\paragraph{Related concepts.}
The lists cover 20 concepts in addition and 26 in multiplication, and Tables~\ref{tab:related}, \ref{tab:relatedadd}, and~\ref{tab:relatedmul} give all of them. Each list holds the concepts from which the standard written algorithm computes its target, and for an operand its digits and for a digit its operand, so that the geometry of an operand is tested against its own digits and that of a digit against its operand. The 62 concepts without a list are the properties of the operands and of the answer, such as parity and magnitude, and the joint concepts.

\begin{table}[ht]
\caption{Related concepts of five core concepts, listed before any experiment.}
\label{tab:related}
\centering
\small
\begin{tabular}{@{}lp{0.24\linewidth}p{0.56\linewidth}@{}}
\toprule
Task & Concept & Related concepts \\
\midrule
Addition & sum $a+b$ & $a$, $b$ \\
Addition & units digit of the answer & $a$, $b$, $a_{\mathrm{units}}$, $b_{\mathrm{units}}$, units column sum, units carry \\
Addition & tens digit of the answer & $a$, $b$, $a_{\mathrm{tens}}$, $b_{\mathrm{tens}}$, tens column sum, units carry, tens carry \\
Multiplication & units carry & units column sum, units partial product \\
Multiplication & units digit of the answer & $a$, $b$, $a_{\mathrm{units}}$, $b_{\mathrm{units}}$, units partial product, units column sum, units carry \\
\bottomrule
\end{tabular}
\end{table}

\begin{table}[ht]
\caption{The 20 lists of related concepts of addition, in the names of Table~\ref{tab:conceptsadd}.}
\label{tab:relatedadd}
\centering
\footnotesize
\begin{tabular}{@{}p{0.40\linewidth}p{0.56\linewidth}@{}}
\toprule
Concept & Related concepts \\
\midrule
\texttt{a} & \texttt{a\_units}, \texttt{a\_tens} \\
\texttt{b} & \texttt{b\_units}, \texttt{b\_tens} \\
\texttt{a\_units}, \texttt{a\_tens} & \texttt{a} \\
\texttt{b\_units}, \texttt{b\_tens} & \texttt{b} \\
\texttt{answer} & \texttt{a}, \texttt{b} \\
\texttt{ans\_units} & \raggedright \texttt{a}, \texttt{b}, \texttt{a\_units}, \texttt{b\_units}, \texttt{column\_sum\_units}, \texttt{carry\_units} \tabularnewline
\texttt{ans\_tens} & \raggedright \texttt{a}, \texttt{b}, \texttt{a\_tens}, \texttt{b\_tens}, \texttt{column\_sum\_tens}, \texttt{carry\_units}, \texttt{carry\_tens} \tabularnewline
\texttt{ans\_hundreds} & \raggedright \texttt{a}, \texttt{b}, \texttt{carry\_tens} \tabularnewline
\texttt{column\_sum\_units} & \texttt{a\_units}, \texttt{b\_units} \\
\texttt{column\_sum\_tens} & \raggedright \texttt{a\_tens}, \texttt{b\_tens} \tabularnewline
\texttt{carry\_units} & \raggedright \texttt{a\_units}, \texttt{b\_units}, \texttt{column\_sum\_units} \tabularnewline
\texttt{carry\_tens} & \raggedright \texttt{a\_tens}, \texttt{b\_tens}, \texttt{column\_sum\_tens}, \texttt{carry\_units} \tabularnewline
\texttt{running\_sum\_units} & \texttt{a\_units}, \texttt{b\_units} \\
\texttt{running\_sum\_tens} & \raggedright \texttt{a\_tens}, \texttt{b\_tens}, \texttt{running\_sum\_units}, \texttt{carry\_units} \tabularnewline
\raggedright \texttt{max\_operand}, \texttt{min\_operand}, \texttt{operand\_diff}, \texttt{operand\_abs\_diff} & \texttt{a}, \texttt{b} \\
\bottomrule
\end{tabular}
\end{table}

\begin{table}[ht]
\caption{The 26 lists of related concepts of multiplication, in the names of Table~\ref{tab:conceptsmul}.}
\label{tab:relatedmul}
\centering
\footnotesize
\begin{tabular}{@{}p{0.40\linewidth}p{0.56\linewidth}@{}}
\toprule
Concept & Related concepts \\
\midrule
\texttt{a} & \texttt{a\_units}, \texttt{a\_tens} \\
\texttt{b} & \texttt{b\_units}, \texttt{b\_tens} \\
\texttt{a\_units}, \texttt{a\_tens} & \texttt{a} \\
\texttt{b\_units}, \texttt{b\_tens} & \texttt{b} \\
\texttt{answer} & \texttt{a}, \texttt{b} \\
\texttt{ans\_units} & \raggedright \texttt{a}, \texttt{b}, \texttt{a\_units}, \texttt{b\_units}, \texttt{partial\_product\_units}, \texttt{column\_sum\_units}, \texttt{carry\_units} \tabularnewline
\texttt{ans\_tens} & \raggedright \texttt{a}, \texttt{b}, \texttt{a\_tens}, \texttt{b\_tens}, \texttt{partial\_product\_a\_units\_b\_tens}, \texttt{partial\_product\_a\_tens\_b\_units}, \texttt{column\_sum\_tens}, \texttt{carry\_units}, \texttt{carry\_tens} \tabularnewline
\texttt{ans\_hundreds} & \raggedright \texttt{a}, \texttt{b}, \texttt{partial\_product\_a\_tens\_b\_tens}, \texttt{column\_sum\_hundreds}, \texttt{carry\_tens} \tabularnewline
\texttt{partial\_product\_units} & \texttt{a\_units}, \texttt{b\_units} \\
\texttt{partial\_product\_a\_units\_b\_tens} & \texttt{a\_units}, \texttt{b\_tens} \\
\texttt{partial\_product\_a\_tens\_b\_units} & \texttt{a\_tens}, \texttt{b\_units} \\
\texttt{partial\_product\_a\_tens\_b\_tens} & \texttt{a\_tens}, \texttt{b\_tens} \\
\texttt{column\_sum\_units} & \texttt{partial\_product\_units} \\
\texttt{column\_sum\_tens} & \raggedright \texttt{partial\_product\_a\_units\_b\_tens}, \texttt{partial\_product\_a\_tens\_b\_units} \tabularnewline
\texttt{column\_sum\_hundreds} & \texttt{partial\_product\_a\_tens\_b\_tens} \\
\texttt{carry\_units} & \raggedright \texttt{column\_sum\_units}, \texttt{partial\_product\_units} \tabularnewline
\texttt{carry\_tens} & \raggedright \texttt{column\_sum\_tens}, \texttt{carry\_units} \tabularnewline
\texttt{running\_sum\_units} & \texttt{partial\_product\_units} \\
\texttt{running\_sum\_tens} & \raggedright \texttt{partial\_product\_a\_units\_b\_tens}, \texttt{partial\_product\_a\_tens\_b\_units}, \texttt{running\_sum\_units}, \texttt{carry\_units} \tabularnewline
\texttt{running\_sum\_hundreds} & \raggedright \texttt{partial\_product\_a\_tens\_b\_tens}, \texttt{running\_sum\_tens}, \texttt{carry\_tens} \tabularnewline
\raggedright \texttt{max\_operand}, \texttt{min\_operand}, \texttt{operand\_diff}, \texttt{operand\_abs\_diff} & \texttt{a}, \texttt{b} \\
\bottomrule
\end{tabular}
\end{table}

\paragraph{All concepts.}
Tables~\ref{tab:conceptsadd} and~\ref{tab:conceptsmul} list all 108 concepts by the names used in our data, in which \texttt{ans} stands for the answer and a double underscore joins the parts of a joint concept.

\begin{table}[ht]
\caption{The 51 concepts of addition.}
\label{tab:conceptsadd}
\centering
\small
\begin{tabular}{@{}p{0.16\linewidth}p{0.80\linewidth}@{}}
\toprule
Group & Concepts \\
\midrule
Core (6) & \raggedright \texttt{a}, \texttt{b}, \texttt{answer}, \texttt{ans\_units}, \texttt{ans\_tens}, \texttt{carry\_units} \tabularnewline
\raggedright Extended single (35) & \raggedright \texttt{a\_eq\_b}, \texttt{a\_is\_zero}, \texttt{a\_magnitude\_tier}, \texttt{a\_num\_digits}, \texttt{a\_parity}, \texttt{a\_tens}, \texttt{a\_units}, \texttt{ans\_ends\_in\_zero}, \texttt{ans\_hundreds}, \texttt{ans\_is\_zero}, \texttt{ans\_magnitude\_tier}, \texttt{ans\_num\_digits}, \texttt{ans\_parity}, \texttt{b\_is\_zero}, \texttt{b\_magnitude\_tier}, \texttt{b\_num\_digits}, \texttt{b\_parity}, \texttt{b\_tens}, \texttt{b\_units}, \texttt{both\_one}, \texttt{both\_zero}, \texttt{carry\_tens}, \texttt{column\_sum\_tens}, \texttt{column\_sum\_units}, \texttt{either\_one}, \texttt{either\_zero}, \texttt{larger\_operand}, \texttt{max\_operand}, \texttt{min\_operand}, \texttt{operand\_abs\_diff}, \texttt{operand\_diff}, \texttt{parity\_match}, \texttt{parity\_xor}, \texttt{running\_sum\_tens}, \texttt{running\_sum\_units} \tabularnewline
Joint (10) & \raggedright \texttt{a\_tens\_\_b\_tens}, \texttt{a\_tens\_\_b\_units}, \texttt{a\_units\_\_b\_tens}, \texttt{a\_units\_\_b\_units}, \texttt{carry\_units\_\_ans\_units}, \texttt{carry\_units\_\_column\_sum\_units}, \texttt{a\_tens\_\_b\_tens\_\_ans\_tens}, \texttt{a\_tens\_\_b\_tens\_\_carry\_units}, \texttt{a\_units\_\_b\_units\_\_ans\_tens}, \texttt{a\_units\_\_b\_units\_\_ans\_units} \tabularnewline
\bottomrule
\end{tabular}
\end{table}

\begin{table}[ht]
\caption{The 57 concepts of multiplication.}
\label{tab:conceptsmul}
\centering
\small
\begin{tabular}{@{}p{0.16\linewidth}p{0.80\linewidth}@{}}
\toprule
Group & Concepts \\
\midrule
Core (6) & \raggedright \texttt{a\_units}, \texttt{a\_tens}, \texttt{b\_units}, \texttt{partial\_product\_units}, \texttt{carry\_units}, \texttt{ans\_units} \tabularnewline
\raggedright Extended single (41) & \raggedright \texttt{a}, \texttt{a\_eq\_b}, \texttt{a\_is\_zero}, \texttt{a\_magnitude\_tier}, \texttt{a\_num\_digits}, \texttt{a\_parity}, \texttt{ans\_ends\_in\_zero}, \texttt{ans\_hundreds}, \texttt{ans\_is\_zero}, \texttt{ans\_magnitude\_tier}, \texttt{ans\_num\_digits}, \texttt{ans\_parity}, \texttt{ans\_tens}, \texttt{answer}, \texttt{b}, \texttt{b\_is\_zero}, \texttt{b\_magnitude\_tier}, \texttt{b\_num\_digits}, \texttt{b\_parity}, \texttt{b\_tens}, \texttt{both\_one}, \texttt{both\_zero}, \texttt{carry\_tens}, \texttt{column\_sum\_hundreds}, \texttt{column\_sum\_tens}, \texttt{column\_sum\_units}, \texttt{either\_one}, \texttt{either\_zero}, \texttt{larger\_operand}, \texttt{max\_operand}, \texttt{min\_operand}, \texttt{operand\_abs\_diff}, \texttt{operand\_diff}, \texttt{parity\_match}, \texttt{parity\_xor}, \texttt{partial\_product\_a\_tens\_b\_tens}, \texttt{partial\_product\_a\_tens\_b\_units}, \texttt{partial\_product\_a\_units\_b\_tens}, \texttt{running\_sum\_hundreds}, \texttt{running\_sum\_tens}, \texttt{running\_sum\_units} \tabularnewline
Joint (10) & \raggedright \texttt{a\_tens\_\_b\_tens}, \texttt{a\_tens\_\_b\_units}, \texttt{a\_units\_\_b\_tens}, \texttt{a\_units\_\_b\_units}, \texttt{carry\_units\_\_ans\_units}, \texttt{carry\_units\_\_column\_sum\_units}, \texttt{a\_tens\_\_b\_tens\_\_ans\_tens}, \texttt{a\_tens\_\_b\_tens\_\_carry\_units}, \texttt{a\_units\_\_b\_units\_\_ans\_tens}, \texttt{a\_units\_\_b\_units\_\_partial\_product\_units} \tabularnewline
\bottomrule
\end{tabular}
\end{table}

\section{Cell Counts and Exclusion Rules}
\label{app:cells}

The cells divide into 2,295 for the 51 addition concepts and 2,565 for the 57 multiplication concepts. Three rules apply in order. First, the floor drops values, and a cell with fewer than two remaining values is removed (Appendix~\ref{app:concepts} gives the values removed). Second, a cell is removed when the permutation test on the singular values keeps no direction, so that $r = 0$. Third, a cell is flagged when its most frequent value has more than three times as many problems as its least frequent value, and flagged cells remain in the analysis. Stage 1 counts are given for the setting without preprocessing, and the later stages pool the three settings (Table~\ref{tab:eligibility}).

\begin{table}[ht]
\caption{Populations of cells, with the setting in which each is counted and the condition that defines it.}
\label{tab:eligibility}
\centering
\small
\begin{tabular}{@{}p{0.30\linewidth}p{0.10\linewidth}p{0.44\linewidth}r@{}}
\toprule
Population & Setting & Condition & Cells \\
\midrule
All cells & all three & every model, task, layer, concept, and setting & 4,860 \\
Candidate cells & one & before any exclusion rule & 1,620 \\
Removed by the floor & off & fewer than two values with 30 problems & 155 \\
Removed by the permutation test & off & no direction kept, $r = 0$ & 20 \\
Decodable cells & off & $r \geq 1$ & 1,445 \\
Cells with a shape fit & pooled & evidence of every candidate shape computed & 1,457 \\
Strongly supported comparisons & pooled & Tier A naming the exact shape & 338 \\
Cells with an ownership verdict & pooled & Tier A or Tier B, a list of related concepts, and $\mQ_{\mathrm{geom}}$ of positive rank & 1,330 \\
Owned cells & pooled & $\Delta < 0$ & 253 \\
Cells with all four claims & pooled & decodable, Tier A, owned, and causally active & 46 \\
\bottomrule
\end{tabular}
\end{table}

\section{Method Details}
\label{app:method}

\paragraph{Probe subspace.}
Let $V$ be the number of values of the concept that remain in the cell, with $N$ and $\bar{\vmu}$ taken over their problems. The matrix decomposed in Stage 1 is $\mC \in \R^{V \times d}$, with rows $\mC_{v,:} = \sqrt{N_v/N}\,(\vmu_v - \bar{\vmu})^\top$. With this weighting, $\mC^\top\mC$ equals the between-class scatter $\frac{1}{N}\sum_v N_v (\vmu_v - \bar{\vmu})(\vmu_v - \bar{\vmu})^\top$, so the right singular vectors of $\mC$ are the directions along which the centroids vary most. Centring forces the last singular value to zero, which leaves at most $V - 1$ directions. Each permutation shuffles $\vy$ while keeping every $N_v$, recomputes the centroids, and records the squared singular values. With $\lambda_j$ the observed squared singular value of rank $j$ and $\tau_j$ the corresponding percentile of its permuted counterpart, the kept rank is
\begin{equation}
r = \max \{ q \in \{0, \dots, V - 1\} \mid \lambda_j > \tau_j \text{ for all } j \leq q \},
\label{eq:rank}
\end{equation}
so that $r = 0$ when the first singular value fails. This step does not whiten the activations. The LDA that follows works on the coordinates inside $\mB$. It regularises the total scatter with Ledoit--Wolf shrinkage \citep{ledoit2004well} when $N/r < 10$, and otherwise adds $10^{-4}$ times the mean eigenvalue to the diagonal. A direction is marked as confirmed, a mark that does not change $\mB$, only if it passes both the rule of Equation~\ref{eq:rank}, applied to the discriminant eigenvalues under 1,000 permutations of $\vy$, and the one-standard-error rule on the accuracy of a nearest-neighbour classifier under five-fold cross-validation. As a consistency check we repeat the LDA in the full $d$-dimensional space with oracle approximating shrinkage \citep[OAS,][]{chen2010shrinkage} and compare the leading directions of the two analyses. An absolute cosine of at least 0.9 counts as agreement, and one from 0.7 to 0.9 as partial agreement. The eigenvalues of the full-space analysis are not interpreted, because with $N/d$ between 0.7 and 2.4 they are inflated by sampling noise. The LDA and this check remove no cell (Appendix~\ref{app:cells}).

\paragraph{Checks on the probe subspace.}
The three checks measure what the probe subspace leaves behind, what it shares with other concepts, and what it keeps of the geometry. For the first, let $\mX^{\mathrm{res}} = \mX(\mI - \mB\mB^\top)$ hold the parts of the activations outside $\mB$, and let $\lambda_1^{\mathrm{res}}$ be the largest eigenvalue of their covariance, computed from the leading 500 components of a randomised SVD. If the residual were noise of the same variance $\sigma_{\mathrm{res}}^2$ in every direction, sampling alone would produce eigenvalues up to the Mar\v{c}enko--Pastur edge \citep{marchenko1967distribution}
\begin{equation}
\lambda_{\mathrm{MP}} = \sigma_{\mathrm{res}}^2 \big(1 + \sqrt{\gamma}\big)^2,
\label{eq:mp}
\end{equation}
where $\gamma$ is the ratio of the dimension of the residual space to $N$ and $\sigma_{\mathrm{res}}^2$ is the trace of the residual covariance divided by that dimension. The check reports the ratio $\lambda_1^{\mathrm{res}} / \lambda_{\mathrm{MP}}$. A ratio above one shows a direction outside $\mB$ whose variance exceeds what noise produces, that is, structure in the residual stream that the probe of this concept does not use. The law describes the limit of large $N$ and large dimension at a fixed $\gamma$, and its edge is a usable reference when $\gamma < 0.7$. Multiplication cells lie above this range, because their correctly answered problems number fewer than the residual dimension, so for them we report the ratio and draw no conclusion from it. For the second check, let $\mB_c$ and $\mB_{c'}$ be the probe bases of two concepts at the same model, layer, and setting. The smallest principal angle between two subspaces is the angle between their closest pair of unit vectors, and it is
\begin{equation}
\theta_1 = \arccos\big( || \mB_c^\top \mB_{c'} ||_2 \big),
\label{eq:angle}
\end{equation}
where $|| \cdot ||_2$ denotes the largest singular value, so that $\theta_1 = 0$ when the subspaces share a direction and $\theta_1 = 90^{\circ}$ when every direction of one is orthogonal to the other. In $d = 4{,}096$ dimensions, two random subspaces of small rank are nearly orthogonal, so under the null hypothesis $\theta_1$ lies close to $90^{\circ}$ at the median rank of nine and lower at the largest ranks, which the pairs drawn at the same ranks reflect. We draw 1,000 pairs of random subspaces of the same ranks, each the orthonormal factor of a Gaussian matrix, and take the fifth percentile of their smallest principal angles. The pair of concepts is superposed when the observed $\theta_1$ lies more than $10^{\circ}$ below this percentile. The margin of $10^{\circ}$ keeps a pair that is only slightly less orthogonal than chance from being marked, and the FDR is controlled over all pairs of concepts. The superposition rate of a model and task is the share of marked pairs, averaged over its cells. For the third check, the squared distance between two activations splits exactly into its parts inside and outside $\mB$,
\begin{equation}
|| \vx_i - \vx_j ||^2 = || \mB\mB^\top(\vx_i - \vx_j) ||^2 + || (\mI - \mB\mB^\top)(\vx_i - \vx_j) ||^2,
\label{eq:pythagoras}
\end{equation}
because $\mB$ has orthonormal columns, and we verify the identity to a relative tolerance of $10^{-6}$ over all pairs. The check then takes the Spearman correlation between the distances $|| \vx_i - \vx_j ||$ and the distances of the projections $|| \mB\mB^\top(\vx_i - \vx_j) ||$ over all pairs of problems and reports its deviation from one. A correlation of one means that the projection keeps the order of every pairwise distance.

\paragraph{Step 2a, period.}
The rows of $\mC$, ordered by value, expressed in the coordinates of $\mB$, and centred, form one sequence per coordinate. We compute the periodogram of every sequence, in which frequency $f$ corresponds to a period of $V/f$ steps of the ordered values. A period is accepted when the strongest frequencies of the two leading coordinates agree within one frequency bin or, failing that, when the power of the strongest frequency, summed over the coordinates, is at least twice that of the second. Otherwise the step declines and reports an inconsistent period. A circle is reported when the accepted frequency in two coordinates holds at least 0.3 of the power of all frequencies and coordinates, when both coordinates pass a permutation test at the 0.01 level, and when the second squared singular value of the matrix of these sequences reaches 0.3 of the first. The last condition rejects a line, which leaks power into the lowest frequency. A helix is reported instead when a linear trend outside the plane of the circle is also significant. Each permutation test compares the observed power with the maximum over all frequencies under shuffles of the order of the values, which accounts for the search over periods, and the FDR over cells is held at 0.05. Cells with fewer than four values are not analysed. The accepted period $T^{\star}$ seeds the candidates $T^{\star}$, $2T^{\star}$, $3T^{\star}$, and $T^{\star}/2$ of Step 2c, and no other period enters Step 2c.

\paragraph{Step 2b, spread.}
The covariance $\hat{\mSigma}_v$ of a value is estimated in the coordinates of $\mB$ as the sample covariance when $N_v/r \geq 10$, with the Ledoit--Wolf estimator when $5 \leq N_v/r < 10$, and with OAS when $N_v/r < 5$. When the values of a cell call for different estimators, all values use the one with the most shrinkage, so that the covariances are comparable. The spread $\sigma_v$ of Equation~\ref{eq:shape} is the square root of $\mathrm{tr}(\hat{\mSigma}_v)/r$, the variance per coordinate averaged over the $r$ coordinates. The step also tests whether the arrangement of the centroids survives once spread is taken into account. It computes the Mahalanobis distance between every pair of centroids under the pooled covariance of the pair, $\tfrac{1}{2}(\hat{\mSigma}_v + \hat{\mSigma}_{v'})$ plus a ridge of $10^{-6}$ times its mean eigenvalue, and takes the Spearman correlation between these distances and the Euclidean ones. The arrangement is confirmed when this correlation is at least 0.85 and the lower end of its 95\% bootstrap interval, from 1,000 resamples, is at least 0.70. A permutation test, held at an FDR of 0.05, guards the result. The outcome of this test is a diagnostic and does not enter the evidence grade.

\begin{table}[ht]
\caption{Candidate shapes of Stage 2. Every feature map also contains the constant 1. The value rescaled linearly to the unit interval is $s$, the angular frequency is $\omega = 2\pi/T$ for a period $T$, the torus has two periods with angular frequencies $\omega_1$ and $\omega_2$, and the radius of the concentric shape is $\eta(v) = 1 + 0.1(s - \tfrac{1}{2}) + 0.05(s - \tfrac{1}{2})^2$.}
\label{tab:shapes}
\centering
\small
\begin{tabular}{@{}lll@{}}
\toprule
Shape & Features $\bm{\phi}_k(v)$ & Description \\
\midrule
$K_0$ generic baseline & $s,\ s^2,\ s^3$ & smooth trend with no periodic feature \\
$K_1$ line & $s$ & values ordered along one direction \\
$K_2$ circle & $\cos\omega v,\ \sin\omega v$ & values repeat with period $T$ \\
$K_3$ open helix & $\cos\omega v,\ \sin\omega v,\ s$ & circle that advances along a line \\
$K_4$ torus & $\cos\omega_1 v,\ \sin\omega_1 v,\ \cos\omega_2 v,\ \sin\omega_2 v$ & product of two circles \\
$K_5$ concentric & $\eta(v)\cos\omega v,\ \eta(v)\sin\omega v,\ \eta(v)$ & circle with a fixed radial drift \\
$K_6$ ribbon & $\cos\omega v,\ \sin\omega v,\ s,\ s\cos\omega v,\ s\sin\omega v$ & helix whose radius changes linearly \\
\bottomrule
\end{tabular}
\end{table}

\begin{figure}[ht]
\centering
\includegraphics[width=\linewidth]{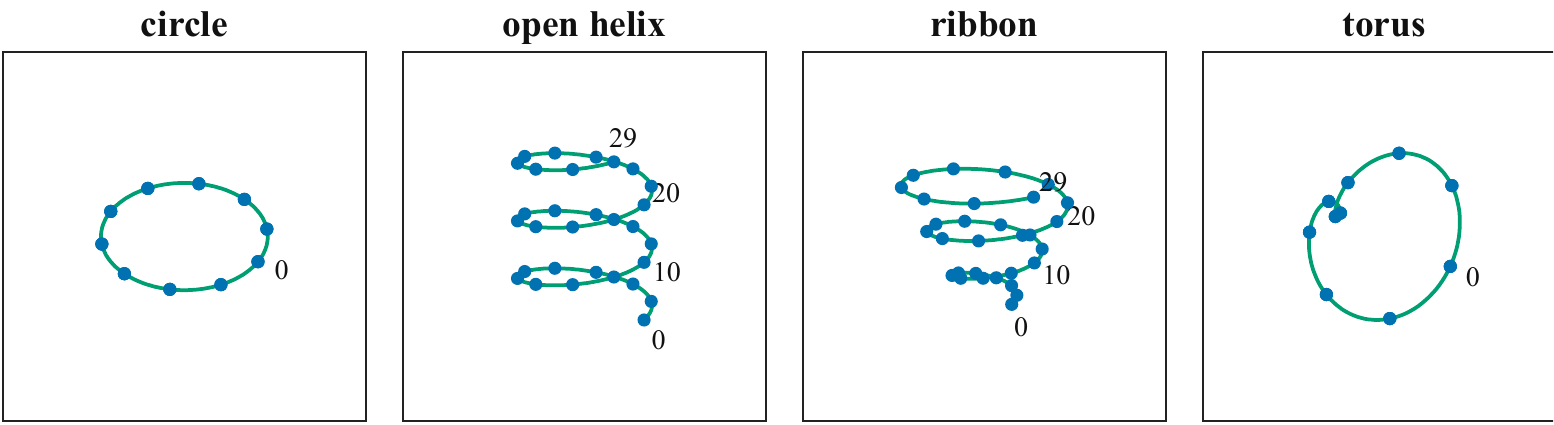}
\caption{The circle, the open helix, the ribbon, and the torus of Table~\ref{tab:shapes} as linear images of their feature maps for the values 0 to 29, with period 10, and periods 10 and 5 for the torus, drawn with one fixed choice of $\mW$ and viewed in three of the coordinates. Dots mark the values, and values ten apart coincide on the closed shapes. The pipeline fits $\mW$, so an observed shape is a rotated, scaled, and possibly flattened image of these curves.}
\label{fig:shapes}
\end{figure}

\paragraph{Step 2c, evidence.}
We stack the coordinates as $\mZ \in \R^{N \times r}$, the features (Table~\ref{tab:shapes}) as $\mPhi_k \in \R^{N \times m_k}$ with rows $\bm{\phi}_k(y_n)^\top$, where $m_k$ counts the features with the constant, and the noise variances as the diagonal matrix $\mD$ with entries $\emD_{n,n} = \sigma_{y_n}^2$. The weights are $\mW \in \R^{m_k \times r}$. Integrating $\mW$ out of Equation~\ref{eq:shape}, conditional on the basis $\mB$ of Stage 1, the noise variances of Step 2b, and the candidate periods of Step 2a, leaves the columns $\mZ_{:,j}$ independent and Gaussian, so that
\begin{equation}
\log Z_k = \sum_{j=1}^{r} \log \gN\big(\mZ_{:,j} ; \vzero,\ \alpha\,\mPhi_k\mPhi_k^\top + \mD\big).
\label{eq:evidence}
\end{equation}
The $N \times N$ covariance is never formed. By the Woodbury identity and the matrix determinant lemma, every term follows from the $m_k \times m_k$ matrix $\mA = \alpha^{-1}\mI + \mPhi_k^\top\mD^{-1}\mPhi_k$, which also gives the posterior mean $\hat{\mW} = \mA^{-1}\mPhi_k^\top\mD^{-1}\mZ$ used in Stage 4, and each fit takes seconds on a CPU. For the ribbon, the two cross features are multiplied by $\sqrt{0.1}$. This gives their weights the scaled prior variance of Section~\ref{sec:method}, and Equation~\ref{eq:evidence} then holds as written with the scaled features. The prior variance $\alpha$ is chosen on a logarithmic grid of 13 points from $10^{-3}$ to $10^{3}$ by maximising the evidence of the best shape. Because this choice is made on the best shape, the stability condition of Tier A repeats the comparison at three fixed values of $\alpha$. The evidence of a periodic shape is the average over the candidate periods of Step 2a, $\log Z_k = \operatorname{logsumexp}_m \log Z_k(T_m) - \log M$ for $M$ candidates, and the torus is averaged in the same way over the six unordered pairs of distinct candidates. For the circle and the open helix, whose period is their only free parameter, a Laplace term is added to each candidate when at least three are evaluated. With the candidates ordered by period, $\ell_m = \log Z_k(T_m)$, and $\delta_m = \tfrac{1}{2}(T_{m+1} - T_{m-1})$, the second difference $h_m = (\ell_{m+1} - 2\ell_m + \ell_{m-1})/\delta_m^2$ estimates the curvature of the log evidence in the period at an interior candidate, and each candidate receives the term $\tfrac{1}{2}\log(2\pi/|h_m|)$ before the average, with $|h_m|$ floored at $10^{-6}$, which also applies to the two end candidates, where no second difference exists. The term is the Laplace estimate of the width of the evidence peak, which the average over a few candidates does not carry. Because Step 2a can decline, and because the baseline and the line have no period, the candidate set does not favour periodic shapes. The line follows the linear account of number representations \citep{zhu2025language}, the circle follows the modular addition solutions of small transformers \citep{liu2022towards, nanda2023progress}, and the open helix is the single-period form of the generalised helix of \citet{kantamneni2025language}. The torus is motivated by residues modulo 10 and the ribbon by the two-digit range, which carries a linear trend together with a periodic one. The concentric shape is a radial variant of the circle, included for completeness. Figure~\ref{fig:shapes} draws the circle, the open helix, the ribbon, and the torus as linear images of their feature maps. Because Equation~\ref{eq:evidence} sums over the coordinates of a cell and each term is the log density of the $N$ values of one coordinate, a gap depends on $N$ and on $r$, and the pipeline compares gaps only within a cell.

\paragraph{Evidence grades.}
The alignment of a shape is $\rho_{\max} = \max(|\rho|, |\rho_{\mathrm{circ}}|)$, where $\rho$ is the Pearson correlation between the concept values and the position of each problem along the fitted linear feature, $\hat{\vw}_s^\top \vz_n / ||\hat{\vw}_s||^2$ for the row $\hat{\vw}_s$ of $\hat{\mW}$ that belongs to the feature $s$, and $\rho_{\mathrm{circ}}$ is the circular correlation between $\omega y_n$ and the angle of each problem in the plane of the fitted cosine and sine features. The held-out error is the mean squared error of the coordinates under five-fold cross-validation, with the weights refitted on the training folds and the period and the noise held, and the error of the best shape may exceed the smallest error of the other shapes by a factor of at most 1.10. The basis of Stage 1, the candidate periods of Step 2a, the noise variances of Step 2b, and $\alpha$ are estimated on all $N$ problems, so the check validates the fitted weights conditional on the estimated representation and hyperparameters and does not test the pipeline as a whole on unseen problems. The best shape is separated from the runner-up when its log evidence, less 1.96 times a standard error from 100 bootstrap resamples of the evidence under a 5\% jitter of the period, leads that of the runner-up by the bar of selection, 5 nats, or 10 nats for six or fewer values, and Tier A then names the exact shape. The stability condition uses the seeds 42, 43, and 44 with $\alpha$ equal to 0.1, 1, and 10. The three families are the shapes with one periodic feature, the circle and the open helix, the shapes with two periodic or radial features, the torus, the concentric shape, and the ribbon, and the line with the baseline. The evidence of a family is the average of the evidence of its members, $\operatorname{logsumexp}$ over the members less the log of their number, and Tier A names only the family when the shape is not separated from its runner-up but the family evidence leads the other families by the bar. Below Tiers A and B, Tier C supports only the dimension of the structure and no shape, and Tier D is the declined verdict. As a guard against multiple testing, every Tier A verdict is also compared with 10,000 permutations of the values, on the statistic $g + 10^{3}\rho_{\max}$, with the FDR controlled at 0.10. Repeating this control at levels of 0.05 and 0.20 leaves the ribbon and the torus as the dominant shapes of Tier A.

\paragraph{Removal of related concepts.}
The bases of the related concepts are stacked and decomposed, and singular vectors whose singular values fall below $10^{-10}$ times the largest are dropped, which removes redundant directions. In the running example $\mR$ spans the subspaces of the operands, of their units digits, of the units partial product, and of the units carry, the units column sum being the same quantity as the partial product (Appendix~\ref{app:concepts}). Because only coordinates inside $\mB$ enter the fit, the removal is computed as $\mZ^{\perp} = \mZ - (\mX - \vone\bar{\vmu}^\top)\,\mR\,(\mR^\top\mB)$, where $\vone$ is the vector of $N$ ones, which needs the small matrix $\mR^\top\mB$ and never a $d \times d$ projection. Row $n$ of $\mZ^{\perp}$ is the vector $\vz_n^{\perp}$ of Section~\ref{sec:method}. The gap $g^{\perp}$ is that of the shape selected in Stage 2, refitted. Let $\mE = \mZ - \mZ^{\perp}$ be the removed part and $\mA_{\Delta} = \mG_0^{-1} - \mG_{\hat{k}}^{-1}$, with $\mG_k = \alpha\mPhi_k\mPhi_k^\top + \mD$. Because the determinant terms of Equation~\ref{eq:evidence} do not depend on the coordinates, and with $\alpha$, the period, and $\mD$ held at their Stage 2 values,
\begin{equation}
\Delta = \tfrac{1}{2}\operatorname{tr}(\mE^\top \mA_{\Delta} \mE) + \operatorname{tr}\big((\mZ^{\perp})^\top \mA_{\Delta} \mE\big).
\label{eq:delta_terms}
\end{equation}
The first term is the advantage of the selected shape over the baseline on the removed part, and the second is the interaction between the removed part and what remains. When the span of $\mR$ lies inside the probe subspace, the removal zeroes whole coordinates, the second term is zero, and $\Delta$ is the first term alone. In general both terms enter, and the verdict is the sign of their sum, with $\Delta = 0$ when the removal takes nothing from $\mB$. Three diagnostics are recorded and do not enter the verdict. They are the share of variance that the removal takes away, whether the family of the selected shape changes, and the ratio of the alignment after and before the removal. A shape whose evidence gap falls is inherited even if it would still be selected, and so is a shape whose coordinates the removal leaves unchanged, with $\Delta = 0$, so the rule is strict at the boundary. The verdict has two limits. The removal can take away directions that belong to the concept itself, which would produce a false inherited verdict, and a related concept missing from the list would produce a false owned verdict.

\paragraph{Interventions.}
The manifold basis is computed by a QR decomposition. It spans the row space of $\hat{\mW}$ expressed in the full activation space, and not the latent parametrisation of the shape. The constant feature is absorbed into the mean, so the bound on the rank is the number of non-constant features of the selected shape, five for the ribbon and four for the torus. Each random control comes from the QR decomposition of a Gaussian matrix with $d$ rows and as many columns as $\mP$, drawn anew for each cell and each type of intervention, with separate seeds for ablation and patching. A random subspace of this rank holds a very small share of the variance of an activation, so the control matches the rank of an intervention and not the variance that its subspace holds. The pairs of donor and recipient are drawn with a fixed seed, and all interventions run in bfloat16 precision in batches of 16. Stages 1 to 3 use all $N$ problems of a cell, and the interventions evaluate at most 256 of them, with the shares of problems answered correctly with and without ablation both read in this pass. The batched bfloat16 pass does not reproduce every answer of the selection pass, so the share without ablation lies below one hundred percent in a few cells, and a problem answered correctly at selection and incorrectly at this baseline can be answered correctly again under ablation, which is the source of the two positive values of Table~\ref{tab:depth}. A cell receives the verdicts of Stages 3 and 4 only if its $\mQ_{\mathrm{geom}}$ also has positive rank.

\begin{table}[ht]
\caption{Fixed thresholds and settings of the pipeline, identical for all cells.}
\label{tab:thresholds}
\centering
\small
\begin{tabular}{@{}p{0.58\linewidth}p{0.37\linewidth}@{}}
\toprule
Quantity & Value \\
\midrule
\multicolumn{2}{@{}l}{\emph{Stage 1}} \\
Minimum number of problems per value & 30 \\
Flag for imbalance, most against least frequent value & more than 3 times \\
Permutations for the rank of the probe subspace, and level & 1,000, 99th percentile \\
Permutations and folds for the LDA directions & 1,000, 5 folds \\
Ledoit--Wolf shrinkage of the total scatter & when $N/r < 10$ \\
Diagonal load otherwise & $10^{-4}$ times the mean eigenvalue \\
Agreement of the two LDAs & absolute cosine $\geq 0.9$, partial from 0.7 \\
Components of the residual decomposition & 500 \\
Reliable range of the Mar\v{c}enko--Pastur comparison & $\gamma < 0.7$ \\
Random pairs, percentile, margin, and FDR for superposition & 1,000, fifth, $10^{\circ}$, 0.05 \\
Tolerance of the distance identity & $10^{-6}$ \\
\midrule
\multicolumn{2}{@{}l}{\emph{Stage 2}} \\
Step 2a shuffles, level, and FDR & 1,000, 0.01, 0.05 \\
Step 2a share of power and ratio of squared singular values & $\geq 0.3$ each \\
Step 2a agreement of periods, and margin of power & one bin, factor 2.0 \\
Step 2a minimum number of values & 4 \\
Step 2b estimator by $N_v/r$ & sample from 10, Ledoit--Wolf from 5, OAS below 5 \\
Step 2b ridge on the pooled covariance & $10^{-6}$ times the mean eigenvalue \\
Step 2b permutations, bootstrap resamples, and FDR & 1,000, 1,000, 0.05 \\
Step 2b confirmed & correlation $\geq 0.85$, lower end $\geq 0.70$ \\
Evidence bar for selection & 5 nats, 10 nats for six or fewer values \\
Alignment, and held-out error factor & $\geq 0.5$, $\leq 1.10$ \\
Seeds and prior variances of the stability condition & 42, 43, 44 and 0.1, 1, 10 \\
Grid for $\alpha$ & 13 points from $10^{-3}$ to $10^{3}$ \\
Prior variance scale of the cross features of the ribbon & 0.1 \\
Permutations, weight of the alignment, and FDR for Tier A & 10,000, $10^{3}$, 0.10 \\
\midrule
\multicolumn{2}{@{}l}{\emph{Stages 3 and 4}} \\
Tolerance when orthonormalising the related concepts & $10^{-10}$ \\
Ownership rule & owned when $\Delta < 0$ \\
Problems per cell for ablation, donor pairs, and random subspaces & at most 256, 100, 5 \\
Seeds of the random subspaces, ablation and patching & 42 and 49 \\
Causally active, and stricter level & 0.5 logits, 1.0 logits \\
\bottomrule
\end{tabular}
\end{table}

\clearpage

\section{Results Details}
\label{app:results}

\paragraph{Stage 1 and Stage 4 by model and task.}
The superposition rate varies little within a model and task, with an SD of at most 0.046 over the cells of a row (Table~\ref{tab:resultsapp}). On the core concepts without preprocessing, 179 of the 180 cells are decodable, and the one cell that is not decodable lies in Pythia addition. Under patching, the logit of the donor answer rises in every model and task, and $\mQ_{\mathrm{geom}}$ keeps from 0.43 to 0.77 of the effect of $\mB$.

\begin{table}[htp]
\caption{Detail by model and task. SD is the standard deviation of the superposition rate over the cells of the row. The ratio $\lambda_1^{\mathrm{res}} / \lambda_{\mathrm{MP}}$ is given as a mean and a maximum over the cells, and the multiplication rows lie outside the range in which the edge is a usable reference (Appendix~\ref{app:method}). Other shapes counts the cells in which the baseline, the circle, or the open helix has the highest evidence. Patching gives the mean causal excess under patching, in logits, on the core concepts over the layers and the three settings.}
\label{tab:resultsapp}
\centering
\small
\setlength{\tabcolsep}{3pt}
\begin{tabular}{@{}llrrrrrrrr@{}}
\toprule
& & \multicolumn{1}{c}{Superposition} & \multicolumn{2}{c}{$\lambda_1^{\mathrm{res}} / \lambda_{\mathrm{MP}}$} & \multicolumn{3}{c}{Other shapes} & \multicolumn{2}{c@{}}{Patching} \\
\cmidrule(lr){3-3} \cmidrule(lr){4-5} \cmidrule(lr){6-8} \cmidrule(l){9-10}
Model & Task & SD & Mean & Maximum & Baseline & Circle & Open helix & $\mB$ & $\mQ_{\mathrm{geom}}$ \\
\midrule
GPT-J & addition & 0.046 & 6.0 & 11.9 & 0 & 0 & 0 & $+2.65$ & $+1.14$ \\
GPT-J & multiplication & 0.014 & 10.0 & 17.3 & 4 & 2 & 1 & $+1.57$ & $+1.19$ \\
Llama & addition & 0.027 & 7.8 & 27.4 & 1 & 0 & 0 & $+1.68$ & $+0.77$ \\
Llama & multiplication & 0.020 & 8.9 & 13.0 & 5 & 0 & 0 & $+0.74$ & $+0.52$ \\
Pythia & addition & 0.046 & 6.9 & 14.1 & 3 & 0 & 0 & $+3.09$ & $+1.62$ \\
Pythia & multiplication & 0.013 & 10.8 & 19.1 & 0 & 2 & 1 & $+1.37$ & $+1.06$ \\
\bottomrule
\end{tabular}
\end{table}

\paragraph{Results by setting.}
Table~\ref{tab:bysetting} gives the shape grades, the ownership verdicts, the retention, and the cells that pass all four criteria under each preprocessing setting. The owned share and the retention change little across the settings, and the cells that pass all four criteria split 16, 14, and 16. Resampling the 122 concepts of the six pairs of model and task 2,000 times, with all layers and settings of a concept drawn together, gives a 95\% interval of 0.54 to 0.67 for the retention, 13\% to 25\% for the owned share, and 37\% to 47\% for the share of causally active cells among the 1,330. Within the extended concepts the shares of causally active cells are 41\% for owned and 40\% for inherited shapes, at layers 14 to 20 they are 69\% and 66\%, and at layers 24 and 28 they are 41\% and 57\%, so the parity of Section~\ref{sec:results} holds where the effects are largest and inherited shapes lead only at the deepest layers.

\begin{table}[ht]
\caption{Results by preprocessing setting over the 1,330 cells with an ownership verdict. Retention is the ratio of the mean accuracy changes under ablation of $\mQ_{\mathrm{geom}}$ and of $\mB$.}
\label{tab:bysetting}
\centering
\small
\begin{tabular}{@{}lrrrr@{}}
\toprule
 & off & answer & norm & all \\
\midrule
Tier A, exact shape & 100 & 118 & 120 & 338 \\
Tier A, family & 31 & 14 & 36 & 81 \\
Tier B & 348 & 245 & 318 & 911 \\
Cells with an ownership verdict & 479 & 377 & 474 & 1,330 \\
Ribbon has the highest evidence & 399 & 314 & 394 & 1,107 \\
Torus has the highest evidence & 78 & 61 & 78 & 217 \\
Owned & 88 & 77 & 88 & 253 \\
Owned share & 18\% & 20\% & 19\% & 19\% \\
Retention & 0.64 & 0.61 & 0.57 & 0.61 \\
Cells that pass all four criteria & 16 & 14 & 16 & 46 \\
\bottomrule
\end{tabular}
\end{table}

\begin{table}[ht]
\caption{Ownership and the cells that pass all four criteria by kind of concept, over the 1,330 cells with a verdict. Functions of the operands are the maximum, the minimum, the difference, and the absolute difference.}
\label{tab:bykind}
\centering
\small
\begin{tabular}{@{}lrrr@{}}
\toprule
Kind of concept & Verdicts & Owned & Pass all four \\
\midrule
Functions of the operands & 252 & 108 & 10 \\
Running sums & 180 & 49 & 7 \\
Operand digits & 298 & 33 & 2 \\
Column sums & 151 & 30 & 7 \\
Operands & 124 & 25 & 17 \\
Partial products & 98 & 5 & 3 \\
Sum of addition & 10 & 3 & 0 \\
Answer digits & 129 & 0 & 0 \\
Carries & 88 & 0 & 0 \\
\midrule
All & 1,330 & 253 & 46 \\
\bottomrule
\end{tabular}
\end{table}

\begin{figure}[ht]
\centering
\includegraphics[width=0.81\linewidth]{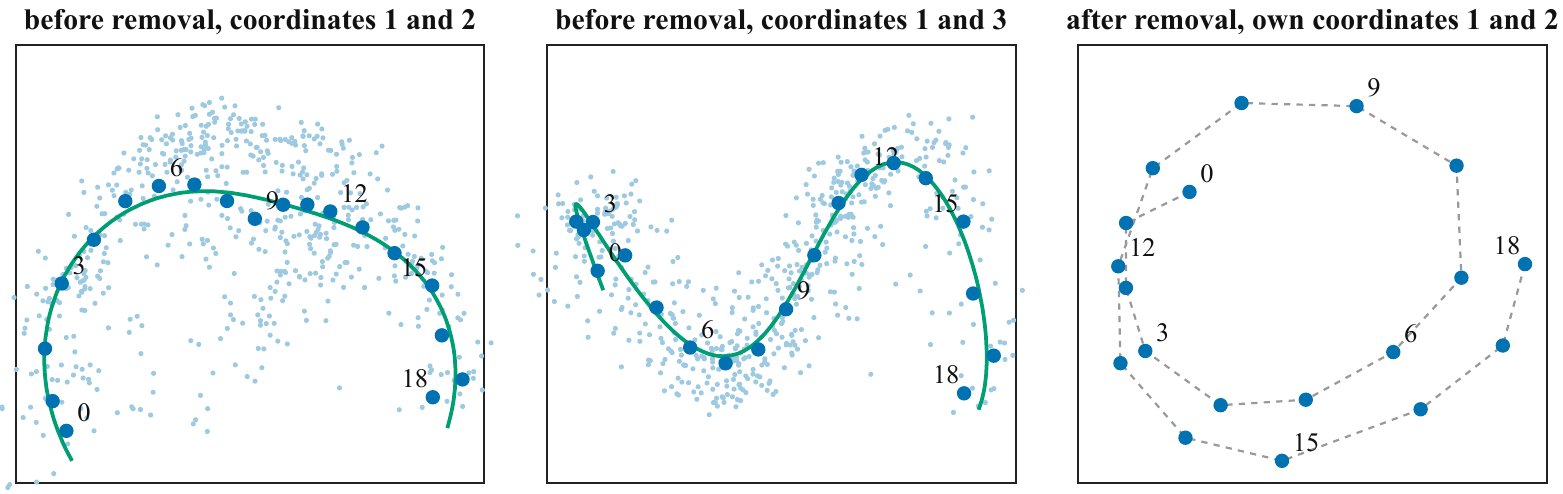}
\caption{The tens column sum of addition in GPT-J at layer 20 without preprocessing, the cell that passes all four criteria. Left and middle, the centroids of its 19 values (dots, every third labelled), a sample of 600 problems (light), and the fitted ribbon (curve), in the three leading principal coordinates of the ribbon inside the manifold basis, which hold 0.72 of the variance of the centroids. Right, the centroids after the subspaces of the two tens digits are removed, in their own two leading coordinates and joined in the order of the values, at 0.49 of the spread before removal. The values wind around twice with a growing radius, the ribbon that Stage 3 refits, which is why the evidence gap grows.}
\label{fig:geometry}
\end{figure}

\paragraph{A cell that passes all four criteria.}
The tens column sum of addition, $a_{\mathrm{tens}} + b_{\mathrm{tens}}$, in GPT-J at layer 20 without preprocessing has 19 values over 8,415 problems, and its related concepts are the two tens digits (Table~\ref{tab:relatedadd}). Its shape is the ribbon in a strongly supported comparison, with an evidence gap of 29,580 nats over the baseline and an alignment of 0.65. Removing the subspaces of the two digits takes 0.70 of the variance of its coordinates, and the gap rises to 55,758 nats, so $\Delta = -26{,}178$ nats and the cell is owned. Ablating its manifold basis lowers the accuracy by 35.5 points against 56.6 under ablation of $\mB$, a retention of 0.63 in this cell, the causal excess is $-2.16$ logits against $-3.04$, and under patching it is $+3.47$ against $+4.53$, with the share answered correctly without ablation in the intervention pass at 99.6\%. Figure~\ref{fig:geometry} shows its centroids on the fitted ribbon and after the removal. We read the cell as a code for the column sum that the two digits it is computed from do not account for, on which the output depends. The running example of Section~\ref{sec:results} gives the inherited outcome that the arithmetic predicts, and this cell gives the owned outcome that it does not.

\paragraph{Shapes of the core concepts.}
Without preprocessing the ribbon has the highest evidence for 24 of the 33 pairs of core concept and model, the torus for 8, and the baseline for 1 (Table~\ref{tab:coreshapes}). The units carry of addition receives no shape verdict in this setting.

\begin{table}[htp]
\caption{Shape with the highest evidence for each core concept and model without preprocessing, as one shape per concept over its five layers.}
\label{tab:coreshapes}
\centering
\small
\setlength{\tabcolsep}{3pt}
\begin{tabular}{@{}lllllll@{}}
\toprule
\multicolumn{7}{@{}l}{Addition} \\
Model & $a$ & $b$ & Answer units & Answer tens & Sum & \\
\midrule
GPT-J & ribbon & ribbon & torus & torus & ribbon & \\
Llama & ribbon & ribbon & ribbon & ribbon & ribbon & \\
Pythia & ribbon & baseline & torus & torus & ribbon & \\
\midrule
\multicolumn{7}{@{}l}{Multiplication} \\
Model & Units of $a$ & Tens of $a$ & Units of $b$ & Partial product & Carry & Answer units \\
\midrule
GPT-J & ribbon & ribbon & ribbon & torus & ribbon & ribbon \\
Llama & ribbon & ribbon & ribbon & ribbon & ribbon & ribbon \\
Pythia & ribbon & ribbon & torus & torus & ribbon & torus \\
\bottomrule
\end{tabular}
\end{table}

\paragraph{Causal effect by layer.}
In the values plotted in Figure~\ref{fig:depth} (Table~\ref{tab:depth}), the change under ablation of $\mQ_{\mathrm{geom}}$ exceeds that under ablation of $\mB$ in magnitude by at most 0.3 points, and the only positive values are $+0.1$ and $+0.2$ points, for GPT-J multiplication at layer 4, from problems that the baseline of the intervention pass answers incorrectly (Appendix~\ref{app:method}).

\begin{table}[!hb]
\caption{Accuracy change under ablation of $\mB$ and of $\mQ_{\mathrm{geom}}$, in percentage points and written as $\mB$ / $\mQ_{\mathrm{geom}}$, by model, task, and layer, over the cells of Figure~\ref{fig:depth}.}
\label{tab:depth}
\centering
\small
\setlength{\tabcolsep}{3pt}
\begin{tabular}{@{}llrrrrr@{}}
\toprule
& & \multicolumn{5}{c@{}}{Layer} \\
\cmidrule(l){3-7}
Model & Task & 4 & 8 & 14 or 16 & 20 or 24 & 24 or 28 \\
\midrule
GPT-J & addition & $-1.9$ / $-1.2$ & $-15.6$ / $-12.5$ & $-53.0$ / $-39.0$ & $-17.1$ / $-9.7$ & $-12.2$ / $-4.8$ \\
GPT-J & multiplication & $+0.1$ / $+0.2$ & $-8.5$ / $-7.4$ & $-14.6$ / $-9.3$ & $-5.9$ / $-5.1$ & $-8.6$ / $-6.2$ \\
Llama & addition & $0.0$ / $0.0$ & $0.0$ / $0.0$ & $-28.3$ / $-12.8$ & $-6.4$ / $-1.4$ & $-15.3$ / $-0.2$ \\
Llama & multiplication & $0.0$ / $0.0$ & $-0.7$ / $-0.7$ & $-3.2$ / $-2.8$ & $-3.7$ / $-2.1$ & $-2.4$ / $-1.2$ \\
Pythia & addition & $-0.5$ / $-0.5$ & $-1.5$ / $-1.8$ & $-47.3$ / $-16.0$ & $-19.1$ / $-9.3$ & $-29.5$ / $-9.3$ \\
Pythia & multiplication & $-0.6$ / $-0.5$ & $-9.4$ / $-8.2$ & $-26.4$ / $-23.4$ & $-6.2$ / $-4.7$ & $-5.6$ / $-4.6$ \\
\bottomrule
\end{tabular}
\end{table}

\end{document}

%% file: math_commands.tex
\usepackage{amsmath,amsfonts,bm}

\def\eqref#1{equation~\ref{#1}}

\def\1{\bm{1}}

\def\vzero{{\bm{0}}}
\def\vone{{\bm{1}}}
\def\vmu{{\bm{\mu}}}

\def\vh{{\bm{h}}}

\def\vw{{\bm{w}}}
\def\vx{{\bm{x}}}
\def\vy{{\bm{y}}}
\def\vz{{\bm{z}}}

\def\mA{{\bm{A}}}
\def\mB{{\bm{B}}}
\def\mC{{\bm{C}}}
\def\mD{{\bm{D}}}
\def\mE{{\bm{E}}}

\def\mG{{\bm{G}}}

\def\mI{{\bm{I}}}

\def\mP{{\bm{P}}}
\def\mQ{{\bm{Q}}}
\def\mR{{\bm{R}}}

\def\mW{{\bm{W}}}
\def\mX{{\bm{X}}}

\def\mZ{{\bm{Z}}}

\def\mPhi{{\bm{\Phi}}}

\def\mSigma{{\bm{\Sigma}}}

\DeclareMathAlphabet{\mathsfit}{\encodingdefault}{\sfdefault}{m}{sl}
\SetMathAlphabet{\mathsfit}{bold}{\encodingdefault}{\sfdefault}{bx}{n}

\def\gN{{\mathcal{N}}}

\def\emD{{D}}

\def\emW{{W}}

\newcommand{\R}{\mathbb{R}}

